%% file: arxiv.tex
\documentclass[letterpaper]{article} 
\usepackage{aaai2027}  
\usepackage[hyphens]{url}  
\usepackage{graphicx} 
\usepackage{natbib}  
\usepackage{caption} 
\usepackage{algorithm}
\usepackage{algorithmic}
\usepackage{amsfonts}
\usepackage{amsmath}
\usepackage{amssymb}

\usepackage{tabularx}
\usepackage{array}
\usepackage[table]{xcolor}
\usepackage{makecell}

\usepackage{pifont}
\newcommand{\cmark}{\ding{51}}

\newcommand{\modelname}{LookAgain}
\definecolor{darkred}{RGB}{139,0,0}
\usepackage{newfloat}
\usepackage{listings}

\usepackage{enumitem}
\usepackage[most]{tcolorbox}

\definecolor{promptgray}{RGB}{245,245,245}
\definecolor{prompttitle}{RGB}{70,70,70}

\lstdefinestyle{promptstyle}{
    basicstyle=\footnotesize\ttfamily,
    numbers=none,
    xleftmargin=0em,
    aboveskip=2pt,
    belowskip=2pt,
    showstringspaces=false,
    tabsize=2,
    breaklines=true,
    breakatwhitespace=false,
    columns=fullflexible,
    keepspaces=true,
    frame=none
}

\newtcolorbox{promptbox}[1]{
    enhanced,
    colback=promptgray,
    colframe=black!70,
    colbacktitle=prompttitle,
    coltitle=white,
    title=#1,
    fonttitle=\bfseries,
    boxrule=0.8pt,
    arc=2mm,
    left=2mm,
    right=2mm,
    top=1mm,
    bottom=1mm
}

\DeclareCaptionStyle{ruled}{labelfont=normalfont,labelsep=colon,strut=off} 
\floatstyle{ruled}
\newfloat{listing}{tb}{lst}{}
\floatname{listing}{Listing}

\usepackage{booktabs}

\nocopyright

\title{LookAgain: Closed-Loop GUI Grounding with Visually Grounded Reflection}
\author{
    Renshan Zhang\textsuperscript{\rm 1},  Haoyang Meng\textsuperscript{\rm 2}, Yixiao He\textsuperscript{\rm 2}, Rui Shao\textsuperscript{\rm 1 3 }\corresponding, April Hua Liu\textsuperscript{\rm 4}, Liqiang Nie\textsuperscript{\rm 1 3 }\corresponding\\
}
\affiliations{
    \textsuperscript{\rm 1}Harbin Institute of Technology, Shenzhen \quad
    \textsuperscript{\rm 2}Beijing University of Posts and Telecommunications \\
    \textsuperscript{\rm 3}Shenzhen Loop Area Institute\quad
    \textsuperscript{\rm 4}Shanghai University of Finance and Economics \\
    
    zhangrenshan@stu.hit.edu.cn \quad rshaojimmy@gmail.com \quad nieliqiang@gmail.com \\
    \textcolor[HTML]{0D47A1}{\url{https://github.com/iLearn-Lab/LookAgain}}
}

\begin{document}

\maketitle

\begin{abstract}
Recent graphical user interface (GUI) grounders have significantly advanced single-shot accuracy on standard benchmarks, yet their performance degrades sharply on small targets, densely packed controls and out-of-distribution interfaces. We attribute this gap to a paradigmatic limitation shared by existing approaches: none of them treats a produced coordinate as a hypothesis to be reflected upon and revised under new visual evidence. This manifests as three coupled issues: 1) \textbf{Lack of post-hoc reflection}. The prediction is frozen at the moment of emission, leaving no internal mechanism to challenge or refine it. 2) \textbf{Visual evidence decoupled from the prediction}. The auxiliary visual evidence is gathered to support the upcoming coordinate rather than to scrutinise the one already committed to. 3) \textbf{Refinement over views, not over predictions}. The iterative zoom-in refines the inspected region instead of inheriting a previous coordinate as a spatial prior to be corrected. In this paper, we propose \textbf{LookAgain}, a closed-loop GUI grounder driven by \textbf{post-prediction visual reflection}. LookAgain reformulates grounding as a multi-turn \emph{predict--look-again--refine} process with two primitives: ``locate" posts a coordinate hypothesis, renders a marker on the image and appends a local patch of the predicted region. It anchors the next reasoning step to the previous prediction as a spatial prior; ``confirm" accepts or reject the hypothesis and terminates the procedure. We train the LookAgain grounder with SFT on constructed reflective trajectories as a cold start, followed by GRPO with terminal grounding correctness as the sole reward.  Extensive experiments show that LookAgain consistently improves performance on both refusal-aware and general GUI grounding benchmarks, achieving state-of-the-art results. Comprehensive ablations further verify the effectiveness of the proposed framework.
\end{abstract}


\input{Sections/sec1_intro_v1}
\input{Sections/sec2_related_works}
\input{Sections/sec3_method_v2}
\input{Sections/sec4_experiments}
\input{Sections/sec5_conclusion}

\bibliography{aaai2027}

\clearpage

\begin{figure*}[!th] 
    \centering
    \includegraphics[width=1\linewidth]{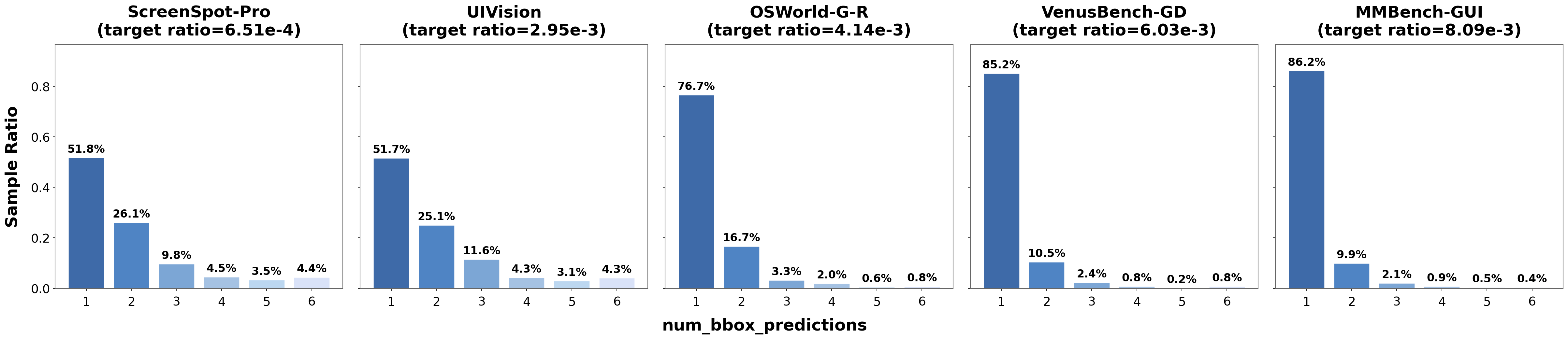}
    \caption{\textbf{Distribution of the number of predicted bounding boxes across different benchmarks.} Each subplot shows the distribution of bounding box prediction counts on one benchmark, with the corresponding target area ratio reported in the title. Consistent with the main discussion in the paper, smaller target area ratios are associated with a larger number of predicted boxes, suggesting that the model tends to produce more predictions to supplement missing details and further refine the localization results.}
    \label{fig:num_bbox_prediction}
\end{figure*}

\section{Implementation Details}

\paragraph{Dataset.}
We build the training data by combining open-source grounding datasets with trajectory synthesis under our proposed closed-loop grounding protocol. Specifically, we manually filter the GroundCUA dataset~\cite{feizi2025groundcua} and retain 20k high-quality grounding samples. We further collect 30k samples from the open-source OS-Atlas dataset~\cite{wu2025atlas}. Based on this mixed dataset, we use Gemini3.1-Pro~\cite{google_gemini_31_pro_preview} to generate multi-turn grounding trajectories following the closed-loop grounding protocol. In particular, the model is prompted to reason about the target region, iteratively make grounding predictions, and verify its own outputs through confirmation turns. We retain only those trajectories whose final prediction correctly localizes the ground-truth bounding box and is further validated by the confirmation step. This process yields about 70k multi-turn grounding trajectories with reflective thinking traces for SFT. The 20k manually filtered high-quality grounding samples also serve as the training data for RL training.

\paragraph{Refusal Data Construction.}
To enable the model to handle inapplicable instructions where the referred element does not exist in the image, we further construct refusal data from public datasets. Inspired by CutPaste~\cite{li2021cutpaste}, we adopt a similar synthesis strategy that takes the grounding samples in OS-Atlas~\cite{wu2025atlas} as the source data to automatically create refusal samples. Specifically, for each source sample, we randomly select one of the four sides surrounding the ground-truth bounding box. We then crop a patch of the same size from that side and paste it onto the original bounding-box region. In this way, the target element referred to by the instruction is overwritten and no longer exists in the image, turning the sample into a valid refusal case. To minimize synthesis artifacts, we further apply feathering to the patch boundaries, producing a smooth transition between the pasted patch and its surroundings. This suppresses visible seams from the manual compositing. Finally, we use Gemini3.1-Pro~\cite{google_gemini_31_pro_preview} to generate multi-turn trajectories for these refusal samples following the same closed-loop grounding protocol. We retain only those whose final \texttt{confirm} step correctly verifies that the target element does not exist. The resulting refusal samples and trajectories are mixed into the training data described above.

\paragraph{Training.}
We implement LookAgain on the Qwen3-VL-8B-Instruct and  Qwen3-VL-32B-Instruct~\cite{bai2025qwen3vl}. For supervised fine-tuning, we use about 70k multi-turn grounding trajectories, among which about 600 are refusal trajectories. We train the model for 1 epoch. The training batch size and learning rate are set to $256$ and $5 \times 10^{-6}$, respectively. For reinforcement learning, we adopt the GRPO~\cite{guo2025deepseek} algorithm and use about 20k training samples, including about 300 refusal samples. We train the model for 3 epochs in this stage, and set the maximum number of interaction rounds to 6 for both RL training and evaluation. The reward coefficients are set to $\alpha = 0.1$ and $\beta = 0.9$. The batch size, learning rate, rollout, and KL loss coefficient are set to $256$, $2 \times 10^{-6}$, $8$, and $1 \times 10^{-3}$, respectively. The system prompt and tool definitions used by LookAgain are shown in Fig.~\ref{fig:system_prompt} and Fig.~\ref{fig:tools_definition}, respectively. We use 16 and 32 H100-80G GPU for training LookAgain 8B and 32B,  respectively. All evaluations are repeated three times to ensure negligible fluctuation in results.

\begin{table}[t]
	\centering
    \setlength{\tabcolsep}{3pt}
    \footnotesize
	\begin{tabular}{lccc}
		\toprule
		\textbf{Model}
		& \textbf{Standard} & \textbf{Refusal} & \textbf{Overall} \\
		\midrule

		\multicolumn{4}{l}{\color{gray}{\textit{$\leq$8B}}} \\
        GTA1-7B \citep{yang2025gta1}     & 74.9 & 0.0 & 67.7 \\
        UI-TARS-1.5-7B \citep{qin2025uitars}     & 71.0 & 0.0 & 64.2 \\
		UI-Venus-7B \citep{gu2025uivenus}      & 65.0 & - & 58.8 \\
        Qwen3-VL-8B$^*$ \citep{bai2025qwen3vl}  & 70.1 & - & 63.4 \\
        \rowcolor{gray!10}
        LookAgain-8B                    & \textbf{83.5} & \textbf{57.4} & \textbf{81.0} \\

        \midrule

        \multicolumn{4}{l}{\color{gray}{\textit{$\geq$30B}}} \\
		GTA1-32B \citep{yang2025gta1}             & 79.8 & 0.0 & 72.2 \\
		OpenCUA-32B \citep{wang2026opencua}       & 76.8 & 7.4 & 70.2 \\
        UI-Venus-72B \citep{gu2025uivenus}           & 77.9 & - & 70.4 \\
        Qwen3-VL-32B$^*$ \citep{bai2025qwen3vl} & 79.9 & - & 72.3 \\
		\rowcolor{gray!10}
		LookAgain-32B                     & \textbf{86.3} & \textbf{50.0} & \textbf{82.8} \\
		\bottomrule
	\end{tabular}
    \caption{
		\textbf{Results of LookAgain on OSWorld-G-Refine.} \textit{Standard} denotes the subset consisting of all non-refusal samples. $^*$ denotes the results evaluated by ourself.
	}
	\label{tab:osg_refine}
\end{table}

\begin{figure*}[t] 
    \centering
    \includegraphics[width=1\linewidth]{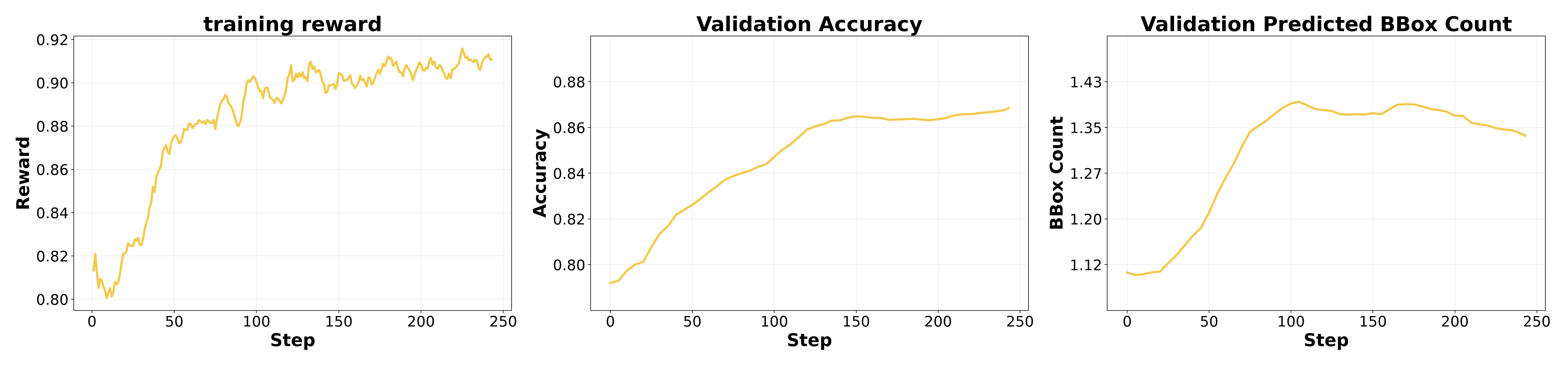}
    \caption{\textbf{Training dynamics of LookAgain.} From left to right, the three plots present the training reward, the validation accuracy, and the average number of predicted bounding boxes on the validation set, respectively. The synchronized increase in validation accuracy and predicted bounding box count suggests that, as training progresses, the model gradually learns to improve grounding by producing more effective refinements through multiple rounds of reflection.}
    \label{fig:training_dynamics}
\end{figure*}

\section{More Experiments}

\paragraph{Results on OSWorld-G-Refine.}
We further include the results on OSWorld-G-Refine~\cite{xie2026osg} in Table~\ref{tab:osg_refine}. Similar to the observations on OSWorld-G~\cite{xie2026osg}, LookAgain achieves strong performance on both the standard and refusal subsets, showing its effectiveness in both regular grounding scenarios and refusal-aware evaluation settings.

\paragraph{Relationship between target size and prediction behavior.} To further validate the relationship between the average number of predicted bounding boxes and the target size, we additionally plot the distribution of the number of bbox predictions under different benchmarks. As shown in Figure~\ref{fig:num_bbox_prediction}, a consistent trend across benchmarks can be observed. When the target area ratio becomes smaller, the model tends to produce more predictions to supplement missing details and further refine the localization results. This observation is aligned with the main discussion in the paper, and further supports that smaller target sizes encourage a more refinement-oriented prediction behavior.

\paragraph{Training dynamics.} We further analyze the training dynamics of the proposed LookAgain. As shown in Figure~\ref{fig:training_dynamics}, as training proceeds, the validation accuracy and the average number of predicted bounding boxes increase simultaneously. This suggests that the model gradually learns to improve grounding through multiple rounds of reflection. Notably, the increase in prediction count is accompanied by better validation performance, indicating that the model is not merely generating more boxes, but producing more effective refinements. We also observe that the predicted bounding box count tends to stabilize and even slightly decrease in the later stage of training, while the validation accuracy continues to improve. This trend suggests that the model gradually shifts from producing more candidate refinements to making more efficient and precise refinement decisions.

\begin{table}[t]
    \centering
    \setlength{\tabcolsep}{2pt}
    \footnotesize
    \begin{tabular}{lccc}
        \toprule
        \textbf{Method} & \textbf{OSG} & \textbf{OSG-R} & \textbf{UIV} \\
        \midrule
        Propose-then-Critic-8B~\cite{wang2026measure} & 59.6 & - & 28.5 \\
        Qwen3-VL-8B w/ MVP~\cite{zhang2026mvp} & - & 72.7 & 31.9 \\
        \rowcolor{gray!10}
        \modelname-8B & \textbf{73.0} & \textbf{81.0} & \textbf{38.8} \\
        \bottomrule
    \end{tabular}
    \caption{
    Comparison with related multi-prediction methods.
    OSG, OSG-R, and UIV denote OSWorld-G, OSWorld-G-Refine, and UI-Vision, respectively.
    }
    \label{tab:related_methods}
\end{table}

\begin{figure*}[!th] 
    \centering
    \includegraphics[width=1\linewidth]{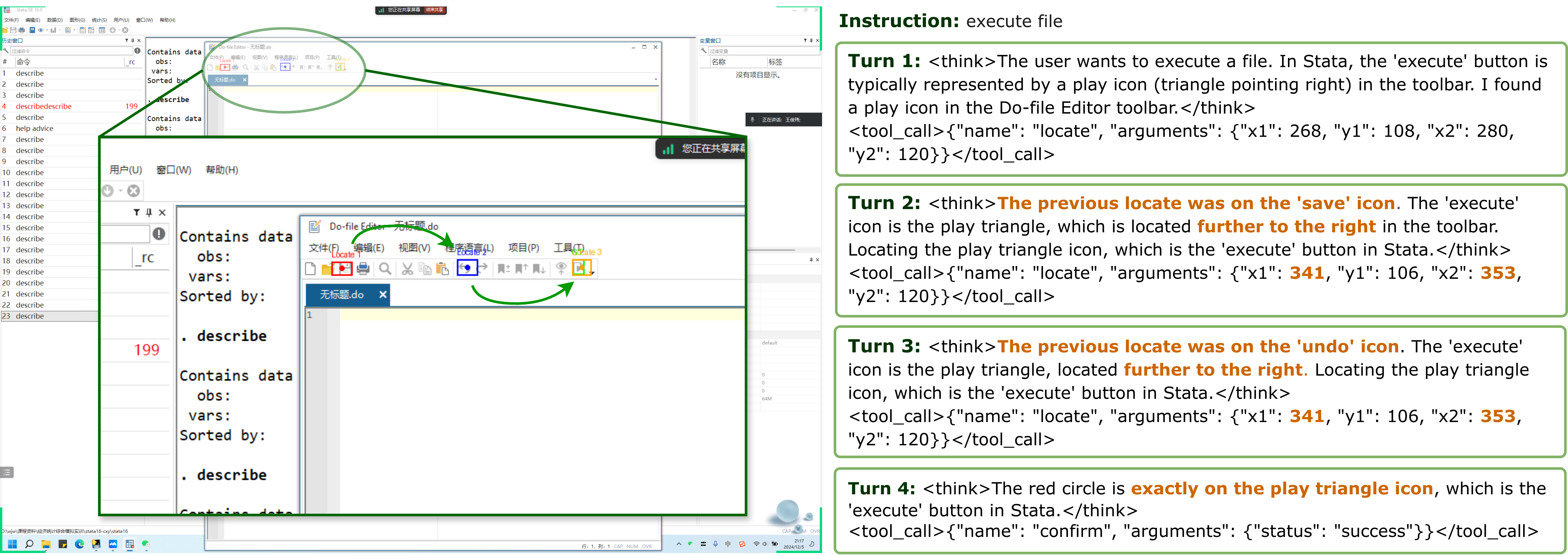}
    \caption{\textbf{A qualitative example illustrating the reflection and refinement capability of LookAgain.} The case shows that LookAgain progressively revises its previous predictions by identifying incorrect toolbar icons and refining the grounding result over multiple turns. This example demonstrates that LookAgain can improve GUI grounding through explicit reflection and iterative refinement.}
    \label{fig:case1}
\end{figure*}

\begin{figure*}[!th]
\centering
\begin{promptbox}{System Prompt}

{\color{black}\bfseries\# Role}

{\color{black}
\begin{itemize}
    \item An agent with strong computer knowledge, designed to locate GUI elements on desktop computer given a screenshot and an instruction.
    \item The user provides tools to help you locate the GUI elements. You are allowed to make one tool call each turn before getting any tool response.
    \item You need to output the coordinate in relative coordinate [0,1000].
    \item Your goal is to make the center of bounding box falls on the target element. As a bonus, you should try your best to make the bounding box cover all interactive area of the target element.
\end{itemize}
}

{\color{black}\bfseries\# Important Notes}

{\color{black}
\begin{itemize}
    \item You should first use \texttt{locate} tool to get bounding box coordinates for the element. The bounding box area will be used to perform mouse action such as clicking, dragging, so it is important to make sure the coordinate is accurate and the bounding box should cover all area that is interactive.
    \item After \texttt{locate} tool is called, you will get an image patch around the bounding box, and you will have a red circle with a marker in its center to indicate where the coordinate is. Then, you will need to use \texttt{confirm} tool to check if the location is correct enough.
    \item If you use \texttt{confirm} with status ``success'', the procedure ends and downstream system gets the coordinates.
    \item It's possible for the user to come up with some instruction that is not applicable (no answer) in this screenshot, you should use \texttt{confirm} tool with status \texttt{failed} and write the reason in \texttt{feedback} field.
    \item If the instruction is applicable and the current location is not correct enough, continue to call \texttt{locate} with a better bounding box.
\end{itemize}
}

\end{promptbox}
\caption{\textbf{System prompt used by LookAgain.}}
\label{fig:system_prompt}
\end{figure*}

\section{Comparison with Related Methods}
\label{sec:related_multi_prediction}

We further compare LookAgain with several recent approaches that also challenge the reliability of single-shot coordinate prediction. Although these methods share the same high-level motivation that GUI grounding can benefit from additional visual evidence beyond one raw prediction, they realize this motivation in different forms. We show that LookAgain provides a more effective closed-loop mechanism for using such evidence through both quantitative comparison and qualitative analysis.

\begin{figure*}[!th]
\centering
\begin{promptbox}{Tools Definition}

{\color{black}\bfseries Tool 1}

{\color{black}
\begin{itemize}[leftmargin=1.2em, itemsep=2pt, topsep=2pt]
    \item \texttt{type}: \texttt{function}
    \item \texttt{function.name}: \texttt{locate}
    \item \texttt{function.description}: locate the element by bounding box
    \item \texttt{function.parameters.type}: \texttt{object}
    \item \texttt{function.parameters.properties}:
    \begin{itemize}[leftmargin=1.5em, itemsep=1pt, topsep=1pt]
        \item \texttt{x1}: \texttt{integer}; bounding box's top left corner x coordinate of the element. Should between 0-1000
        \item \texttt{y1}: \texttt{integer}; bounding box's top left corner y coordinate of the element. Should between 0-1000
        \item \texttt{x2}: \texttt{integer}; bounding box's bottom right corner x coordinate of the element. Should between 0-1000
        \item \texttt{y2}: \texttt{integer}; bounding box's bottom right corner y coordinate of the element. Should between 0-1000
    \end{itemize}
    \item \texttt{function.parameters.required}: [\texttt{x1}, \texttt{y1}, \texttt{x2}, \texttt{y2}]
\end{itemize}
}

\vspace{0.4em}

{\color{black}\bfseries Tool 2}

{\color{black}
\begin{itemize}[leftmargin=1.2em, itemsep=2pt, topsep=2pt]
    \item \texttt{type}: \texttt{function}
    \item \texttt{function.name}: \texttt{confirm}
    \item \texttt{function.description}: Confirm the location of the element or confirm that the instruction is not applicable
    \item \texttt{function.parameters.type}: \texttt{object}
    \item \texttt{function.parameters.properties}:
    \begin{itemize}[leftmargin=1.5em, itemsep=1pt, topsep=1pt]
        \item \texttt{status}: \texttt{string}; enum = [\texttt{success}, \texttt{failed}]; Status of the confirmation. Should be one of 'success', 'failed'
        \item \texttt{feedback}: \texttt{string}; Feedback of the confirmation. Should have this field when status=failed
    \end{itemize}
    \item \texttt{function.parameters.required}: [\texttt{status}]
\end{itemize}
}

\end{promptbox}
\caption{\textbf{Tools definition used by the grounding agent.}}
\label{fig:tools_definition}
\end{figure*}

\paragraph{Comparison with MVP.}
MVP~\cite{zhang2026mvp} introduces a multi-view proposal strategy for GUI grounding. It constructs multiple views of the same interface and performs grounding independently on each view. The resulting coordinates are then aggregated through clustering to produce the final prediction. This design improves grounding robustness by collecting diverse coordinate hypotheses. However, the role of these multiple predictions is mainly to provide statistical consensus. Since each hypothesis is produced independently, the model does not explicitly inspect a previous prediction, identify why it may be wrong, or refine it based on visual feedback. In this sense, MVP provides multiple proposals but does not establish a reflect-and-refine loop. In contrast, LookAgain turns each prediction into a visible hypothesis and uses the corresponding local evidence to decide whether to confirm or revise it. This makes the correction process prediction-conditioned rather than merely proposal-aggregated. As shown in Table~\ref{tab:related_methods}, LookAgain-8B outperforms MVP across both benchmarks, suggesting that grounding reflection in the model's own previous action is more effective than aggregating independently generated proposals.

\paragraph{Comparison with Propose-then-Critic.}
Propose-then-Critic~\cite{wang2026measure} further introduces a critic into the grounding process. It first generates a set of candidate predictions. These candidates are then rendered with visual identifiers, and a critic selects the most plausible one as the final answer. Compared with MVP, it introduces a more explicit judgment step and can be viewed as an initial form of reflection. Nevertheless, the critic mainly performs discrimination among pre-defined candidates. Its final answer is still bounded by the coverage and quality of the candidate set produced in the first stage. LookAgain instead formulates GUI grounding as an iterative closed-loop process. Rather than selecting from fixed candidates, it marks its previous prediction on the screen and observes the corresponding local context. When the marked location is inconsistent with the instruction, LookAgain can make a new prediction based on this visual feedback. This enables LookAgain to actively revise its own prediction trajectory rather than only judge a pre-defined candidate set. As shown in Table~\ref{tab:related_methods}, LookAgain-8B outperforms Propose-then-Critic-8B on both benchmarks, indicating that a critic-style selection step is less effective than iterative, visually grounded revision.

\begin{figure*}[!t] 
    \centering
    \includegraphics[width=0.925\linewidth]{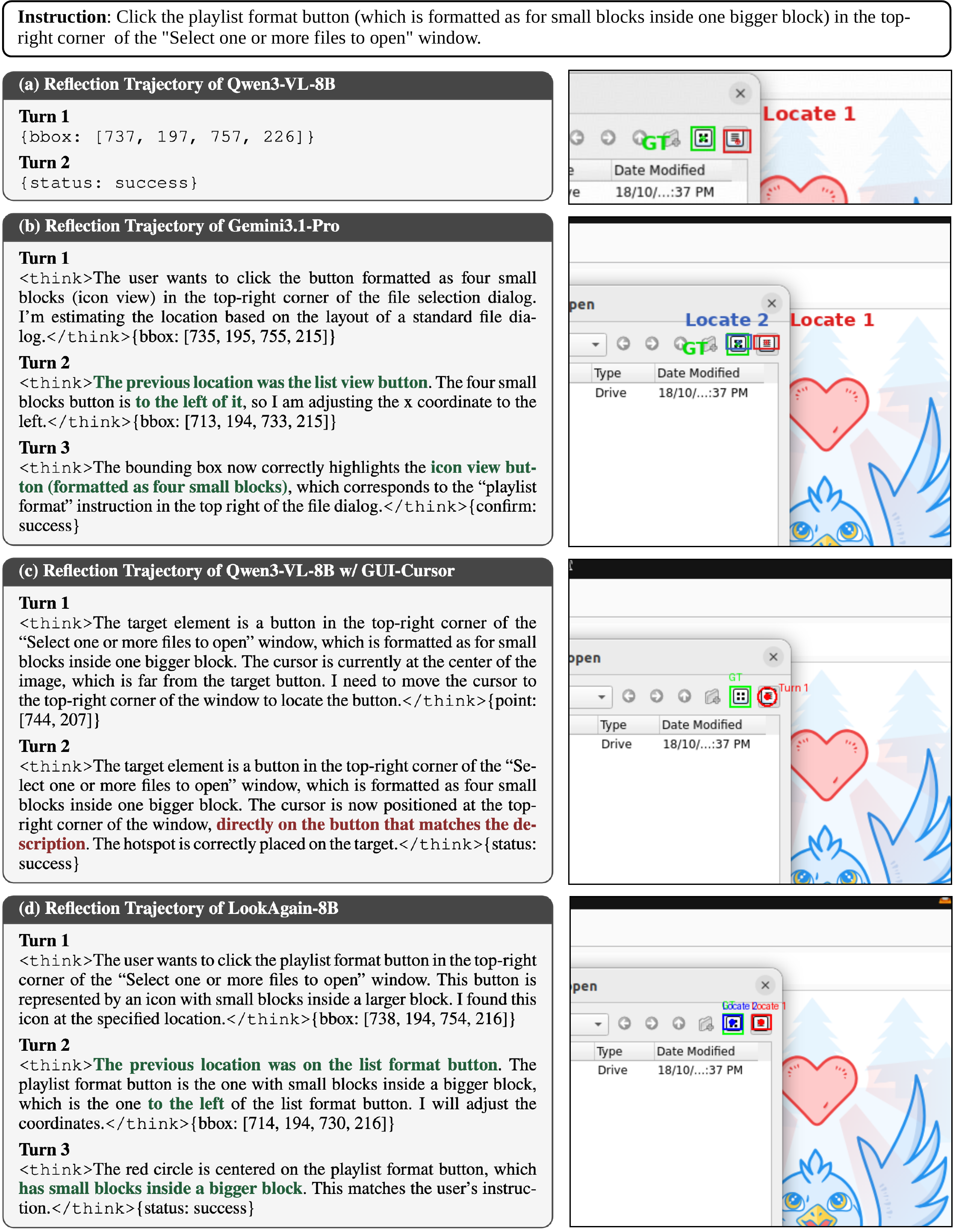}
    \caption{\textbf{Reflection trajectory comparison of four grounders.} Turn-by-turn trajectories of (a) Qwen3-VL-8B-Instruct, (b) Gemini3.1-Pro w/ closed-loop protocol, (c) the GUI-Cursor variant, and (d) LookAgain-8B, showing that only Gemini3.1-Pro and LookAgain-8B perform genuine reflect-and-refine, whereas the others commit an unreflective or hallucinated prediction.}
    \label{fig:traj_compare}
\end{figure*}

\section{Qualitative Results}

\paragraph{Case Study} We provide a qualitative example in Figure~\ref{fig:case1} to illustrate the reflection and refinement capability of LookAgain. In this case, the model is asked to execute a file in the Stata interface. Instead of identifying the correct target in a single step, LookAgain progressively revises its prediction by recognizing that the previously selected locations correspond to incorrect toolbar icons, such as the save or undo button, and then shifts its attention to the correct play icon. This example shows that LookAgain can explicitly reflect on its previous predictions and refine the grounding result through iterative correction, leading to a more accurate final decision.

\paragraph{Reflection Trajectory Comparison.}
To complement the quantitative comparison of direct closed-loop grounding in the paper, we provide reflection traces in Fig.~\ref{fig:traj_compare}, where four grounders are given the same grounding sample. We visualize their turn-by-turn trajectories on the corresponding annotated screenshots.As shown in Fig.~\ref{fig:traj_compare}(a), Qwen3-VL-8B-Instruct~\cite{bai2025qwen3vl} issues a single \textsc{locate} that lands on the adjacent list-view button and then immediately emits \textsc{confirm}(\texttt{success}) without any intervening reasoning. Since no \verb|<think>| block is produced, the model never inspects the marker-annotated patch returned for its own hypothesis, and the initial error is committed as the final prediction. This exactly matches the failure mode analyzed in paper. It is in the absence of explicit reflection, the closed loop degenerates into a one-shot prediction, which is why directly transferring the strategy to Qwen3-VL-8B-Instruct fails to bring gains.

In contrast, as shown in Fig.~\ref{fig:traj_compare}(b), Gemini3.1-Pro~\cite{google_gemini_31_pro_preview} exhibits the intended reflection behavior. Its first \textsc{locate} also falls on the list-view button, but in Turn~2 it explicitly recognizes the mistake and reasons about the correct spatial relation, shifting the coordinate onto the icon-view target before confirming in Turn~3. The Turn-1 hypothesis is treated as an addressable spatial prior to be refined rather than a frozen output, which is precisely the mechanism behind its substantial improvement.

However, as shown in Fig.~\ref{fig:traj_compare}(c), directly training Qwen3-VL-8B-Instruct with GUI-Cursor~\cite{zhao2025learning} style method elicits only superficial reflection. The model emits a \verb|<think>| block, yet its Turn-2 reasoning hallucinates success, and it confirms without performing any actual refinement. This corroborates our observation that the reflect-and-refine pattern cannot be reliably induced by direct reinforcement learning alone, and explains the limited gain of the GUI-Cursor variant.

Finally, as shown in Fig.~\ref{fig:traj_compare}(d), LookAgain-8B reproduces the reflective pattern of the much stronger Gemini3.1-Pro within an 8B backbone. Its first \textsc{locate} lands on the list-view button, Turn~2 correctly diagnoses the error and moves the coordinate left, and Turn~3 verifies that the red ring is centered on the four-block icon before issuing \textsc{confirm}. This side-by-side comparison demonstrates that our two-stage pipeline instills genuine, visually grounded reflection—rather than the mimicry seen in Fig.~\ref{fig:traj_compare}(c), thereby accounting for the significant improvement reported.

\section{Broader Impact and Potential Risk}
\label{appendix_impact}

Our approach employs readily available LLMs, which means it inherently shares some of their limitations. This includes the potential for hallucinating ungrounded text or producing biased results. We recommend conducting a thorough investigation into its safety and fairness for the intended use before applying it in practice.


\end{document}

%% file: Sections/sec1_intro_v1.tex
\section{Introduction}
Graphical user interface (GUI) grounding, which maps a natural-language instruction to a precise on-screen coordinate, has become a fundamental capability for screen-operating agents~\cite{qin2025uitars,lyu2026personalalign,zhou2025hiconagent,chen2025less}. Early works such as SeeClick~\cite{cheng2024seeclick}, OS-Atlas~\cite{wu2025atlas} and UI-TARS~\cite{qin2025uitars} have substantially advanced single-shot accuracy on standard benchmarks. However, their performance degrades sharply small targets and densely packed layouts. Such grounding errors remain the dominant source of cascading failures in downstream agents.

\begin{figure}[t] 
    \centering
    \includegraphics[width=1\linewidth]{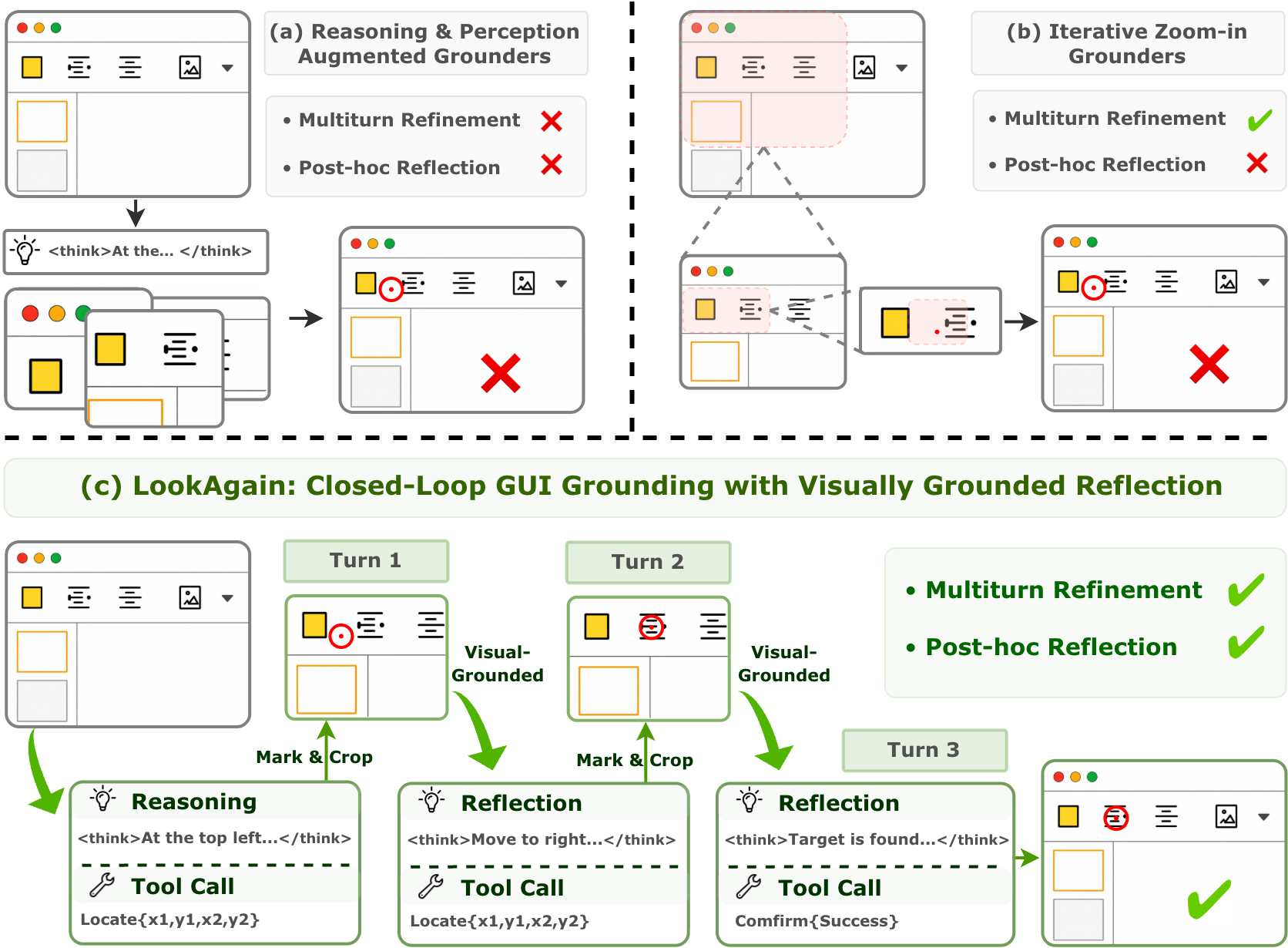}
    \caption{Comparison between existing GUI grounding paradigms and LookAgain. Existing methods either improve reasoning and perception before prediction or refine visual views across turns, but do not perform post-hoc reflection on a predicted coordinate. LookAgain instead turns grounding into a closed-loop predict–look-again–refine process through marker-based cropping and visually grounded reflection.}
    \label{fig:intro}
\end{figure}

To narrow this gap, recent efforts strengthen GUI grounders along several directions. Reasoning-augmented grounders strengthen the textual chain of thought that precedes coordinate emission through tailored training objectives~\cite{chen2026uiins} or reward shaping~\cite{zhou2026guig1,tang2026guig2}. Attention-guided perception grounders enrich a single forward pass with auxiliary attention signals~\cite{wu2026guiactor} or multi-view evidence~\cite{zhang2026mvp} to mitigate the difficulty of perceiving small or cluttered elements. Iterative zoom-in grounders~\cite{luo2025regionfocus,wu2025dimo} further allow the model to consume additional cropped views across multiple turns before committing to a final coordinate. These works further improve grounding accuracy. However, they still lack a protocol that allows the model to reflect on previously produced coordinates, ultimately bounding their effectiveness.

We attribute the remaining gap to a limitation shared by existing grounders. Although recent methods strengthen reasoning, perception, or iterative refinement through zoom-in rounds, none of them treats a produced coordinate as a hypothesis to be reflected upon and revises under new visual evidence. Specifically, we identify three coupled issues:

\textbf{1) Lack of post-hoc reflection.} Existing grounders commit to a coordinate without ever treating it as a hypothesis to be reflected. The prediction is therefore frozen at the moment of emission, leaving no internal mechanism through which the model can challenge or update its own decision.

\textbf{2) Visual evidence decoupled from the prediction.} Auxiliary views and attention maps are gathered to support the upcoming prediction. They are not used to scrutinise it afterwards. The visual evidence the model relies on is therefore never aligned with the hypothesis it ultimately commits to.

\textbf{3) Refinement over views, not over predictions.} In iterative zoom-in, each round refines the region under inspection rather than a concrete coordinate. The model never inherits a previous prediction as a spatial prior. Independent re-samples thus yield diminishing returns on difficult cases.

To resolve these issues, we revisit GUI grounding from the perspective of what happens \emph{after} a prediction is made, rather than how to better perceive or reason before it. Self-refinement in language modelling~\cite{shinn2023reflexion} suggests that critiquing one's own output is more effective than thinking longer up front.
However, purely textual self-critique cannot tell whether a predicted pixel actually hosts the requested element. We therefore propose \emph{visually grounded reflection}.
The model first emits a coordinate hypothesis. It is then presented with a high-resolution view of that exact location and decides whether to commit or to refine. Grounding is thus reformulated from a one-shot regression into a closed \textbf{\emph{predict--look-again--refine}} process.

We instantiate this paradigm as a multi-turn tool-use protocol with two primitives. A ``locate" call posts a coordinate hypothesis on the screen. A marker is rendered at the box center on the original image. A high-resolution patch around the predicted region is then cropped and appended to the model's context. A ``confirm" call accepts the current hypothesis or rejects the instruction as inapplicable, and terminates the procedure. At each turn the model first reasons in a thinking block and then chooses to issue another locate for refinement or to terminate via confirm. The reasoning is therefore anchored to what the model just claimed to see. This design enforces three properties: (i) every reasoning step is conditioned on the model's own previous prediction; (ii) every newly observed view is summoned by, and tied to, that prediction; and (iii) refinement inherits the spatial prior of the previous round instead of restarting from scratch.

To realise this paradigm, we build a two-stage training pipeline. We first construct multi-turn reflective grounding trajectories, in which each sample interleaves reasoning, ``locate" hypotheses, marker-annotated high-resolution patches and a terminating ``confirm" call. These trajectories serve as a cold-start corpus for supervised fine-tuning (SFT), teaching the model the tool-use format and the basic predict--look-again--refine behaviour. We then further optimise the model with Group Relative Policy Optimisation (GRPO), using a simple outcome reward of grounding correctness at the terminating step. This stage encourages the model to issue refinements only when they improve the final coordinate and to commit when the current hypothesis is already accurate.

We evaluate LookAgain on extensive GUI grounding benchmarks. LookAgain-8B achieves 73.0 overall accuracy on OSWorld-G, improving over the base Qwen3-VL-8B by 21.7 points. It further reaches 84.5 on MMBench-L2-GUI and brings especially large gains on challenging benchmarks such as UIVision (+15.5) and ScreenSpot-Pro (+13.1). These results show that revisiting a predicted coordinate under newly grounded visual evidence leads to more reliable grounding decisions. We also conduct extensive ablations, which consistently validate the effectiveness of the proposed closed-loop reflection paradigm and training recipe.

We summarise our main contributions as follows:
\begin{itemize}
    \item We propose \textbf{LookAgain}, a closed-loop GUI grounding paradigm driven by visually grounded reflection. It turns grounding into a multi-turn \emph{predict-look-again-refine} loop where \verb|locate| and \verb|confirm| primitives anchor reflection to the model's previous prediction.
    \item We design a two-stage SFT+GRPO recipe that cold-starts the reflective behaviour on constructed multi-turn trajectories and refines it with terminal grounding correctness as the sole reward.
    \item Extensive experiments demonstrate that LookAgain consistently improves performance of both refusal-aware and general GUI grounding. Comprehensive ablations further validate the effectiveness of the proposed framework.
\end{itemize}

%% file: Sections/sec2_related_works.tex
\begin{figure*}[t] 
    \centering
    \includegraphics[width=1\linewidth]{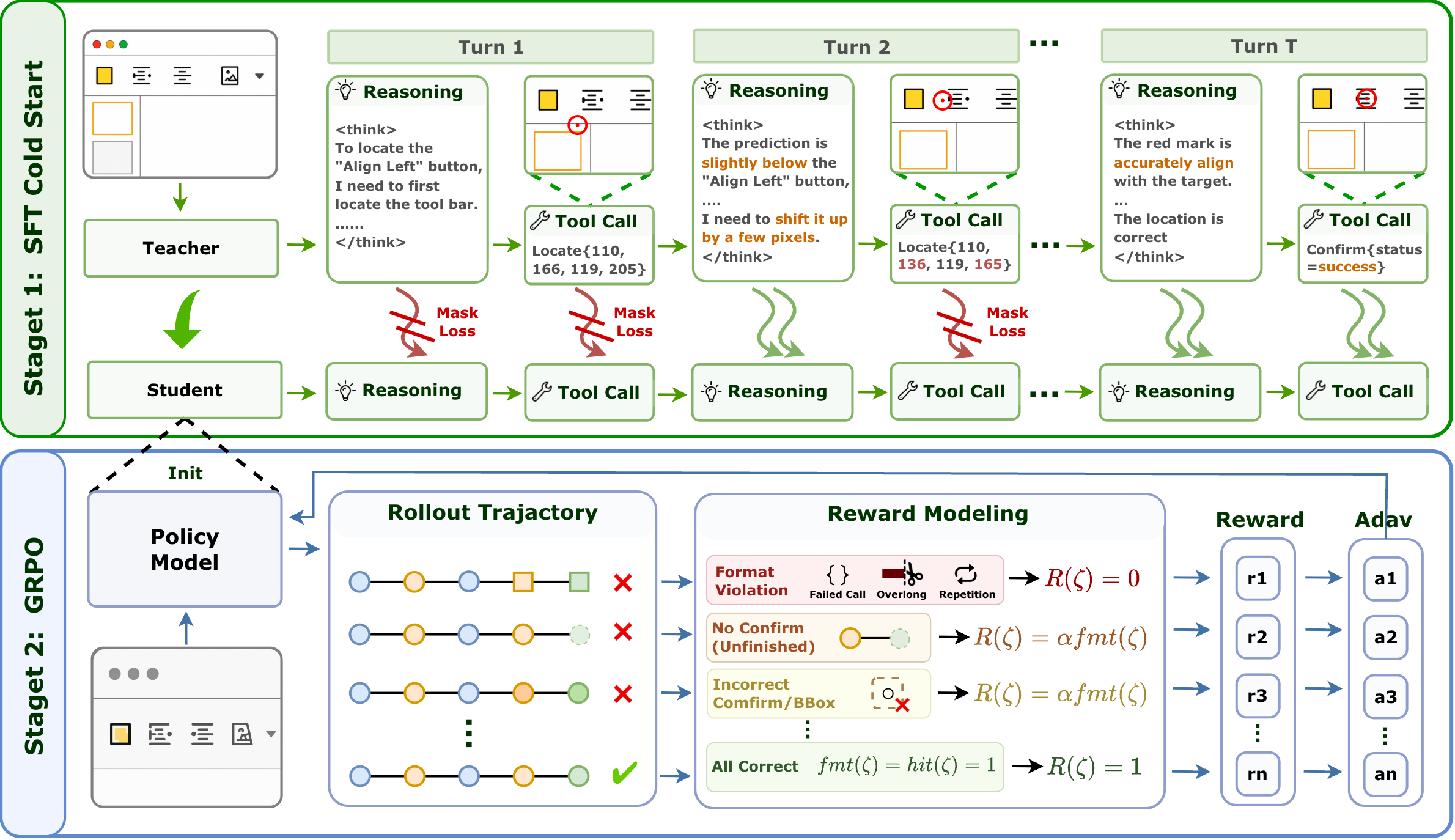}
    \caption{Overall training framework of LookAgain. The model is first cold-started by supervised fine-tuning on teacher-generated reflective trajectories with selective loss masking. It is then further optimized by GRPO with simple rule-based rewards over rollout trajectories.}
    \label{fig:arch}
\end{figure*}

\section{Related Works}

\paragraph{GUI Grounding.}
Recent advances in multimodal understanding\cite{shao2019multi,shao2023detecting,shao2024detecting}, multimodal large language models (MLLMs)~\cite{zhang2025falcon,li2025lion} and vision-language-action models~\cite{li2026cogvla,li2026semanticvla,hu2026behavior,li2025optimus} have demonstrated the potential of grounding visual understanding into action decisions. Building on this progress, GUI grounding aims to map a natural-language instruction to an on-screen coordinate. Existing grounders predominantly cast this as a one-shot regression, in which a MLLM takes the screenshot and instruction and directly emits the coordinate in a single forward pass~\cite{cheng2024seeclick,gou2025navigating,wu2025atlas,yang2025aria,qin2025uitars}. Representative efforts include SeeClick~\cite{cheng2024seeclick} with grounding-oriented pretraining and OS-Atlas~\cite{wu2025atlas} scaling to larger cross-platform GUI corpora. However, this forces the model to commit a coordinate without explicit deliberation. It also restricts the evidence to one global observation that is rarely sufficient for small elements, leading to substantial accuracy degradation~\cite{li2025screenspot,nayak2025ui}.

\paragraph{Reasoning-Augmented Grounding.}
To address the lack of explicit deliberation, a line of reasoning-augmented grounders strengthens the chain-of-thought reasoning before coordinate emission~\cite{lu2026ui,luo2025gui,liu2025infigui,zhou2026guig1,tang2026guig2,chen2026uiins}. For example, UI-R1~\cite{lu2026ui} and GUI-R1~\cite{luo2025gui} introduce R1-style~\cite{guo2025deepseek} reasoning to enhance the deliberation prior to grounding. GUI-G1~\cite{zhou2026guig1} and GUI-G2~\cite{tang2026guig2} further design geometry-aware reward shaping, providing a stronger learning signal. UI-INS~\cite{chen2026uiins} instead reshapes the grounding reasoning by recasting the instruction as multi-perspective rationales.
These methods substantially improve grounding accuracy, yet their reasoning is still produced before the coordinate is emitted and thus brings in no new visual evidence to verify the previous prediction.

\paragraph{Perception-Augmented Grounding.}
To address the perception bottleneck on small elements, another line of work augments the visual evidence available to the grounder~\cite{wu2026guiactor,chen2026v2p,zhang2026mvp}. For example, GUI-Actor~\cite{wu2026guiactor} and V2P~\cite{chen2026v2p} read out attention maps to localise relevant patches before coordinate emission. MVP~\cite{zhang2026mvp} instead runs independent inference on several attention-guided crops and aggregates the predictions via spatial clustering. These methods enrich the evidence supporting the upcoming coordinate, yet never turn it back to scrutinise one already committed to.

\paragraph{Iterative Zoom-in.}
To overcome the limitations of single-pass grounding, iterative zoom-in methods have emerged that progressively narrow the inspected region across multiple rounds. RegionFocus~\cite{luo2025regionfocus} performs visual test-time scaling by repeatedly selecting salient regions guided by the model's own reasoning. DiMo-GUI~\cite{wu2025dimo} further leverages modality-aware visual reasoning to decide where to crop at each step. However, these multi-round procedures only refine the inspected region without explicitly reflecting on the previous prediction. In this work, we treat each prediction as a hypothesis to be revisited, iterating over coordinates rather than over views and turning naive zoom-in into a closed predict--look-again--refine loop.

%% file: Sections/sec3_method_v2.tex
\section{Method}
\label{sec:method}

We present \textbf{LookAgain}, a closed-loop GUI grounder that reformulates grounding as a multi-turn \emph{predict--look-again--refine} process. We first formalise the protocol that drive visually grounded reflection, then describe the two-stage training pipeline that instils the reflective behaviour.

\subsection{Closed-Loop Grounding via Visually Grounded Reflection}
\label{sec:method:loop}

\paragraph{Problem reformulation.} Given a screenshot $I$ and a natural-language instruction $q$, conventional grounders learn a one-shot mapping $f_\theta:(I,q)\!\mapsto\!\hat{p}\in[0,1000]^2$ that emits a single normalised click point. We instead model grounding as a sequential decision process $\pi_\theta:(I,q,h_{<t})\!\mapsto\!a_t$, where at each turn $t$ the policy inspects the dialogue history $h_{<t}$ (the original screenshot, prior thoughts, prior prediction and associated visual patches) and emits a thought $\tau_t$ followed by exactly one tool call $a_t\!\in\!\{\textsc{locate},\textsc{confirm}\}$. The episode terminates as soon as a \textsc{confirm} is issued, and the centre of the last \textsc{locate} call is taken as the final prediction.

\paragraph{The \textsc{locate} primitive.} A call $\textsc{locate}(x_1,y_1,x_2,y_2)$ with $x_i,y_i\!\in\![0,1000]$ posts a coordinate hypothesis on the original screenshot. The tool then performs three operations to create visual evidence for further reflection: (i) it computes the bounding-box centre $(c_x,c_y)$ and overlays a visual marker consisting of a small solid red dot at $(c_x,c_y)$ enclosed by a hollow red ring, producing an annotated copy $I'_t$ of the original image; (ii) it crops $I'_t$ into $512\!\times\!512$ patch $P_t$ centred at $(c_x,c_y)$; (iii) it appends the $P_t$ to the dialogue history as the tool response. In this way, every patch the model later inspects is the direct visual consequence of its own previous prediction. The hypothesis therefore becomes the addressable subject of the next reasoning step rather than a frozen, unreviewable output.

\paragraph{The \textsc{confirm} primitive.} A call $\textsc{confirm}(\text{status})$ commits the location or refuses the instruction and terminates the episode. \textsc{status}=\texttt{success} commits the centre of the most recent \textsc{locate} as the final prediction. \textsc{status}=\texttt{failed}, accompanied by a free-text \texttt{feedback}, declares the instruction infeasible on the given screenshot. It allows the model to abstain rather than emit an arbitrary coordinate.

\paragraph{Closed Grounding Loop.} The two primitives are composed into a closed loop that runs for as many turns as the model deems necessary. At each turn the policy is required to first emit a \verb|<think>...</think>| block $\tau$ and then exactly one tool call $a$, conditioned on the running context $h_{<t}=(I,q,\tau_1,a_1,P_1,\ldots,\tau_{t-1},a_{t-1},P_{t-1})$. The thought is thus anchored either (i) to the marker-annotated patch from its previous hypothesis, enabling \emph{post-hoc} reflection on whether the red dot actually landed on the target element, or (ii) for the first turn, to the raw screenshot alone. The loop accumulates evidence rather than overwriting it. All previously rendered patches $P_t$ are retained in $h_{<t}$, so the model can compare consecutive hypotheses against each other and against the original screenshot. A \textsc{locate} call in turn $t\!+\!1$ thus inherits turn $t$'s coordinate as an explicit spatial prior to be refined, rather than restarting the search from scratch, and the loop closes when the model is confident enough to issue a \textsc{confirm} or when the maximum turn limit is reached.

\begin{figure}[t] 
    \centering
    \includegraphics[width=1\linewidth]{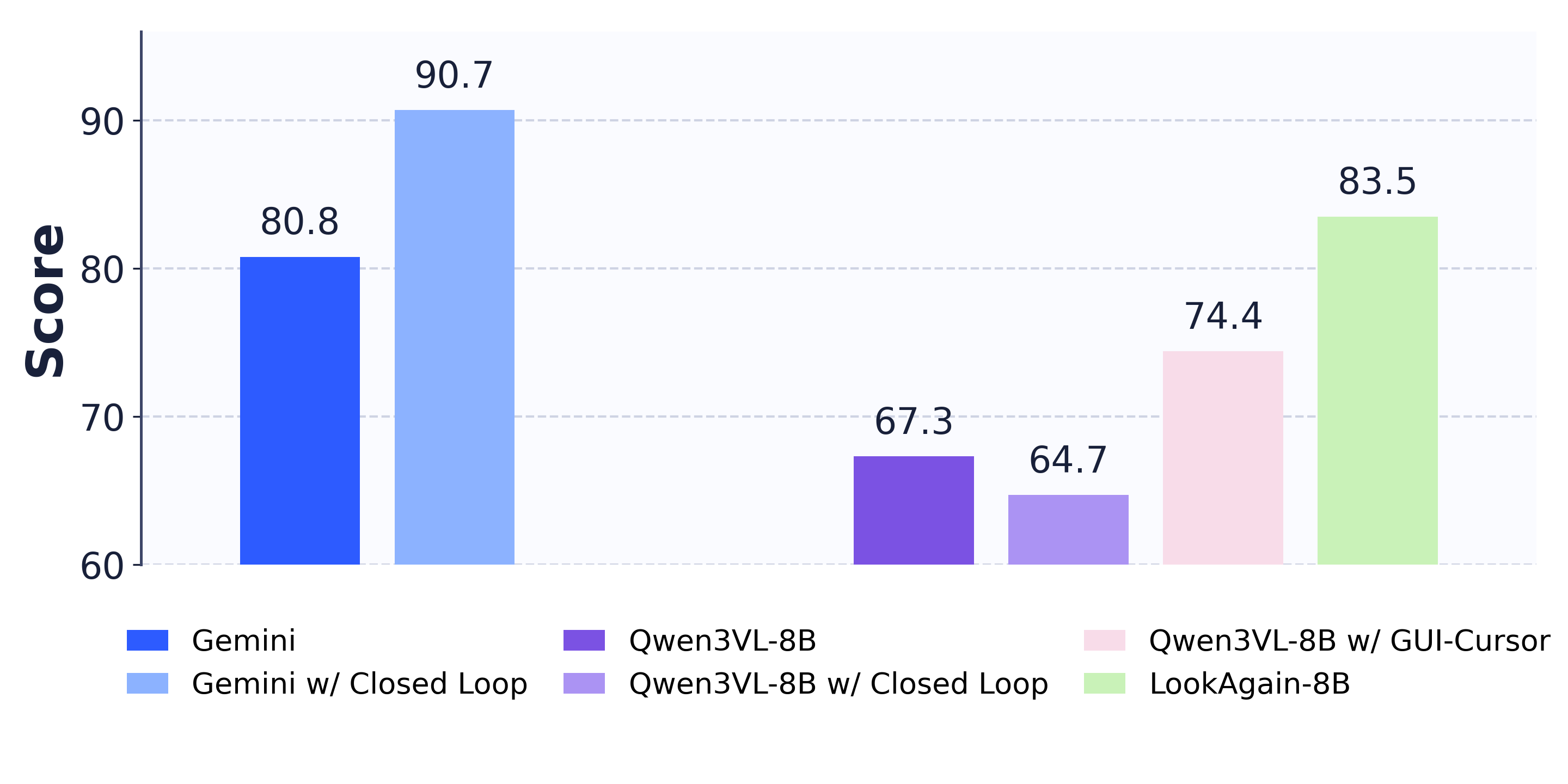}
    \caption{Direct application of closed-loop grounding across different backbones on OSWorld-G-Refine. Benefiting from effective reflection and refinement, Gemini3.1-Pro achieves a substantial improvement. In contrast, Qwen3-VL-8B-Instruct degrades due to the lack of effective reflection, and such reflective behavior cannot be effectively elicited by GUI-Cursor style training, which only gains limited improvement. LookAgain learns this reflection capability through our two-stage pipeline and achieves a significant improvement.}
    \label{fig:acc_closed_loop}
\end{figure}

\paragraph{Analysis of Direct Closed-Loop Grounding.} We first examine whether the proposed closed-loop grounding strategy can be directly transferred to different foundation models. As shown in Fig.~\ref{fig:acc_closed_loop}, applying the closed-loop grounding procedure to Gemini3.1-Pro~\cite{google_gemini_31_pro_preview} brings a clear improvement on OSWorld-G-Refine~\cite{xie2026osg}. However, directly applying the same strategy to Qwen3-VL-8B-Instruct~\cite{bai2025qwen3vl} does not yield similar gains. Instead, the performance drops due to incorrectly revising originally correct predictions. We hypothesize that this degradation stems from the absence of explicit reflection and refinement. Unlike Gemini, which can naturally produce reflective reasoning before refining its prediction, Qwen3-VL-8B-Instruct~\cite{bai2025qwen3vl} directly predicts the grounding result without such an intermediate reasoning process. Moreover, this reflection and refinement pattern is not easily elicited by direct reinforcement learning alone. To verify this, we train the closed-loop grounding policy following GUI-Cursor~\cite{zhao2025learning}. Although the model learns a certain degree of reflective behavior, the gain remains limited. We also provide qualitative comparisons of their reflection traces in the \textbf{Appendix}. These observations motivate our two-stage training pipeline.


\subsection{Two-Stage Training: SFT Cold Start and GRPO}
\label{sec:method:training}

We train LookAgain in two stages. The first stage instils the predict--look-again--refine tool-use behavior on synthetic reflective trajectories distilled from a teacher model. The second stage uses reinforcement learning to push the model towards trajectories whose terminal commitment is actually correct, while suppressing degenerate behaviours such as malformed tool calls and locate repetition.

\subsubsection{Stage 1: Supervised Cold Start on Reflective Trajectories}
\label{sec:method:sft}

\paragraph{Trajectory collection.} We collect $(I,q,\mathbf{b}^{\star})$ grounding triples from publicly available GUI grounding corpora, where $\mathbf{b}^{\star}$ is the ground-truth bounding box. For each retained triple we obtain a multi-turn reflective trajectory by rolling out a strong teacher VLM~\cite{google_gemini_31_pro_preview} under the afordmentioned tool-use protocol. At every turn the teacher first produces a \verb|<think>| block and then issues a \textsc{locate} or \textsc{confirm} tool call, while the environment executes the call and returns the marker-annotated patch or terminate. We retain a trajectory only if it terminates with \textsc{confirm}(\texttt{success}) and the centre of the last \textsc{locate} falls inside $\mathbf{b}^{\star}$. This filtering yields a corpus of self-consistent reflective trajectories in which every \verb|<think>| block is conditioned on the visual evidence rendered by the preceding hypothesis.

\paragraph{SFT objective.} Let $\mathcal{D}_{\text{SFT}}=\{(I^{(i)},q^{(i)},\zeta^{(i)})\}_{i=1}^{N}$ denote the corpus, where each trajectory $\zeta^{(i)}=(\tau^{(i)}_1,a^{(i)}_1,P^{(i)}_1,\dots,\tau^{(i)}_{T_i},a^{(i)}_{T_i})$ is composed of thoughts $\tau_t$, tool calls $a_t$ and marker-annotated patches $P_t$, with $a_t$ being a \textsc{locate} for $t<T_i$ and a \textsc{confirm} at $t=T_i$. Patch tokens are always masked from the loss. For the assistant-emitted tokens we apply a \emph{turn-selective} mask that reflects which parts of a teacher trajectory carry a credible learning signal. Concretely, let $T^{\star}_i\!=\!T_i\!-\!1$ denote the index of the final \textsc{locate} (the one whose centre fell inside $\mathbf{b}^{\star}$ by construction) and $\tau^{(i)}_{T_i}$, $a^{(i)}_{T_i}$ the closing thought and \textsc{confirm}. We define the trainable token set $\mathcal{S}(\zeta^{(i)})$ as follows:
\begin{itemize}
  \item If $T^{\star}_i\!=\!1$ (the very first \textsc{locate} was already correct), $\mathcal{S}$ contains every assistant token, i.e.\ $\tau_1,a_1,\tau_2,a_2$ are all supervised.
  \item Otherwise, the first turn's $\tau^{(i)}_1$ and $a^{(i)}_1$ are masked out, since both the reflection-free opening thought and the wrong opening hypothesis carry no useful signal. For intermediate turns $1\!<\!t\!<\!T^{\star}_i$, only the \textsc{locate} call $a^{(i)}_t$ is masked while the preceding thought $\tau^{(i)}_t$ remains supervised, since this thought is a genuine post-hoc reflection on the previous marker-annotated patch even though the refined coordinate still misses. The final correcting turn $t\!=\!T^{\star}_i$ and the closing \textsc{confirm} turn $t\!=\!T_i$ are fully supervised.
\end{itemize}
The training objective is then
\begin{equation}
\small
\mathcal{L}_{\text{SFT}}(\theta) = -\!\sum_{i=1}^{N}\sum_{u\in\mathcal{S}(\zeta^{(i)})} \log \pi_\theta\!\big(u \,\big|\, I^{(i)},q^{(i)},h^{(i)}_{<u}\big).
\end{equation}
This selective scheme cold-starts the model with three behaviours at once: emitting an accurate first hypothesis when the screenshot is unambiguous (from one-shot-correct trajectories), reflecting critically on a marker-annotated patch of one's own wrong hypothesis (from supervised intermediate thoughts), and committing the refined coordinate via \textsc{confirm} (from fully supervised final turns), without ever imitating a coordinate that the teacher itself got wrong.

\subsubsection{Stage 2: Reinforcement Learning with GRPO}
\label{sec:method:grpo}

To directly optimise the final grounding outcome instead of merely imitating teacher trajectories, we further train the model with Group Relative Policy Optimisation (GRPO) under a rule-based reward.

\paragraph{Trajectory-level reward.} For each rollout $\zeta$ on input $(I,q,\mathbf{b}^{\star})$, let $\hat{\mathbf{b}}$ denote the bbox of the last \textsc{locate} in $\zeta$ and $s\in\{\texttt{success},\texttt{failed},\varnothing\}$ the status of the terminating \textsc{confirm}, where $\varnothing$ means no \textsc{confirm} is issued. We define
\begin{align}
\mathrm{hit}(\zeta) &= \mathbb{1}[s\!=\!\texttt{success}]\cdot \mathbb{1}[\mathrm{c}(\hat{\mathbf{b}})\!\in\!\mathbf{b}^{\star}], \\
\mathrm{fmt}(\zeta) &= \mathbb{1}[\neg\mathrm{Violation}(\zeta)],
\end{align}
where $\mathrm{c}(\cdot)$ returns the bbox centre.    Specifically, if the current sample is a refusal sample, we instead define
\[
\mathrm{hit}(\zeta)=\mathbb{1}[s=\texttt{failed}].
\]
The $\mathrm{Violation}(\zeta)$ is true if any of the following holds: (i) the rollout has a malformed tool call; (ii) the same \textsc{locate} bbox is emitted more than $K$ times in $\zeta$; (iii) the trajectory was truncated by the response-length cap. The final reward is
\begin{equation}
R(\zeta) = \alpha\cdot\mathrm{fmt}(\zeta) + \beta\cdot\mathrm{hit}(\zeta).
\label{eq:reward}
\end{equation}
Note that the hit term is gated on a successful \textsc{confirm}, i.e. an intermediate \textsc{locate} that happens to fall on the target but is never committed earns no hit credit. This forces the policy to treat \textsc{confirm} as a genuine commitment rather than an afterthought. It also eliminates a degenerate rollout pattern in which an early trajectory is accidentally correct on the first \textsc{locate} but then collapses into post-correct rambling. Moreover, we zero out both reward terms when any of the three violations occurs. This helps GRPO to see a clean negative signal for runaway loops, instead of still earning the hit reward whenever a violating trajectory happens to contain a GT-aligned \textsc{locate}.

\begin{table*}[!t]
	\centering
	\footnotesize
    \setlength{\tabcolsep}{5pt}
	\begin{tabular}{l ccc ccc}
		\toprule
		\textbf{Model}
		& \multicolumn{3}{c}{\textbf{OSWorld-G}}
		& \multicolumn{3}{c}{\textbf{VenusBench-GD   }} \\
		\cmidrule(lr){2-4} \cmidrule(lr){5-7}
		& \textbf{Standard} & \textbf{Refusal} & \textbf{Overall}
		& \textbf{Standard} & \textbf{Refusal} & \textbf{Overall} \\
		\midrule

		\multicolumn{7}{l}{\color{gray}{\textit{$\leq$8B}}} \\
        GTA1-7B \citep{yang2025gta1}     & 60.9 & 0.0 & 55.1 & 54.3 & 0.0 & 46.4 \\
        UI-TARS-1.5-7B \citep{qin2025uitars}     & 58.4 & 0.0 & 52.8 & 47.6 & 0.2 & 40.7 \\
		OpenCUA-7B \citep{wang2026opencua}        & -    & - & 55.3 & 56.4 & 0.0 & 48.2 \\
		UI-Venus-7B \citep{gu2025uivenus}      & 60.4 & - & 54.6 & 57.4 & 0.0 & 49.0 \\
        GUI-Cursor \cite{zhao2025learning} & 64.1 & 0.0 & 58.0 & - & - & - \\
        Holo1.5-7B \citep{hai2025holo15modelfamily}           & - & - & - & 59.7 & 0.0 & 51.0 \\
        VISTA-8B \citep{qiu2026vista}  & - & - & 62.4 & - & - & - \\
        VISTA-8B w/ MVP \citep{zhang2026mvp}  & - & - & 63.1 & - & - & - \\
        Qwen3-VL-8B$^*$ \citep{bai2025qwen3vl}  & 56.7 & - & 51.3 & 62.3 & 57.3 & 61.5 \\
        
        \rowcolor{gray!10}
        LookAgain-8B                    & \textbf{74.7} & \textbf{57.4} & \textbf{73.0} & \textbf{69.7} & \textbf{70.6} & \textbf{69.8} \\

        \midrule

        \multicolumn{7}{l}{\color{gray}{\textit{$\geq$30B}}} \\
		GTA1-32B \citep{yang2025gta1}             & 72.1 & 0.0 & 65.2 & 68.9 & 0.0 & 58.8 \\
		OpenCUA-32B \citep{wang2026opencua}       & - & - & - & 58.6 & 0.0 & 50.1 \\
        UI-Venus-72B \citep{gu2025uivenus}           & 68.8 & - & 62.2 & 73.5 & 51.3 & 70.2 \\
        Holo1.5-72B \citep{hai2025holo15modelfamily}     & - & - & - & 73.7 & 0.0 & 62.9 \\
        
        Qwen3-VL-32B$^*$ \citep{bai2025qwen3vl} & 67.0 & - & 60.6 & 65.6 & 55.1 & 64.1 \\
		\rowcolor{gray!10}
		LookAgain-32B                     & \textbf{78.4} & \textbf{50.0} & \textbf{75.7} & 71.3 & \textbf{69.6} & \textbf{71.1} \\
		\bottomrule
	\end{tabular}
    \caption{
		\textbf{Results of LookAgain on OSWorld-G and Venus-GD.} \textit{standard} denotes the subset consisting of all non-refusal samples. $^*$ denotes the results evaluated in this work.
	}
    \label{tab:main_refusal_results}
\end{table*}

\paragraph{GRPO objective.} For each prompt $(I,q,\mathbf{b}^{\star})$ we sample a group of $G$ rollouts $\{\zeta_g\}_{g=1}^{G}$ from the current policy under the defined multi-turn protocol, compute $\{R(\zeta_g)\}$ via Eq.~\eqref{eq:reward}, and form group-normalised advantages
\begin{equation}
A_g = \frac{R(\zeta_g)-\mu_R}{\sigma_R+\varepsilon},
\end{equation}
where $\mu_R$ and $\sigma_R$ are the within-group mean and standard deviation of $R$. The policy is updated with the GRPO algorithm
\begin{equation}
\small
\mathcal{J}_{\text{GRPO}}(\theta) = \mathbb{E}\!\!\left[ \frac{1}{G}\!\sum_{g=1}^{G}\! \frac{1}{|\mathcal{U}_g|}\!\sum_{u\in\mathcal{U}_g}\! \big( \mathcal{M}_{g,u}(\theta) - \lambda\,\hat{\mathbb{D}}_{\mathrm{KL}}(u) \big) \right],
\end{equation}
with the clipped surrogate
\begin{equation}
\small
\mathcal{M}_{g,u}(\theta) = \min\!\big( r_{g,u}(\theta)\,A_g,\; \mathrm{clip}(r_{g,u}(\theta),1\!-\!\epsilon,1\!+\!\epsilon)\,A_g \big),
\end{equation}
where $r_{g,u}(\theta)\!=\!\pi_\theta(u|\cdot)/\pi_{\theta_{\text{old}}}(u|\cdot)$ is the per-token importance ratio, $\mathcal{U}_g$ is the set of assistant-emitted tokens in $\zeta_g$, and $\epsilon,\lambda$ are the clip range and KL coefficient.


%% file: Sections/sec4_experiments.tex
\section{Experiments}

We implement LookAgain on the Qwen3-VL~\cite{bai2025qwen3vl} series. For more training details, please refer to \textbf{Appendix}.



\begin{table}[!t]
	\centering
	\footnotesize
	\setlength{\tabcolsep}{1.75pt}
	\begin{tabular}{l c c c}
		\toprule
		\textbf{Model} & \textbf{SSPro} & \textbf{\makecell{UIVision}} & \textbf{\makecell{MMB-\\L2-GUI}} \\
		\midrule
		\multicolumn{4}{l}{\color{gray}{\textit{$\leq$8B}}} \\
		UI-TARS-1.5-7B \citep{qin2025uitars}     & 35.7     & 22.3     & 64.3    \\
		OpenCUA-7B \citep{wang2026opencua}        & 50.0     & 29.7     & -       \\
		GTA1-7B \citep{yang2025gta1}              & 50.1     & -     & 78.5       \\
		UI-Venus-7B \citep{gu2025uivenus}      & 50.8     & 26.5     & 79.9    \\
		GUI-Owl-7B \citep{ye2025mobile}  & 54.9     & -     & 80.5    \\
		Holo2-8B \citep{hai2025holo2modelfamily}            & 58.9     & 35.1     & 84.5    \\
        Qwen3-VL-8B$^*$ \citep{bai2025qwen3vl}  & 47.1     & 23.3     & 79.3    \\
        \rowcolor{gray!10}
        LookAgain-8B                      & \textbf{60.2}    & \textbf{38.8}     & \textbf{84.5}     \\
		
        \midrule
		
        \multicolumn{4}{l}{\color{gray}{\textit{$\geq$30B}}}   \\
		
		OpenCUA-32B \citep{wang2026opencua}       & 55.3     & 33.3     & -           \\
		GUI-Owl-32B \citep{ye2025mobile} & 58.0     & -     & 83.0        \\
		GTA1-32B \citep{yang2025gta1}             & 63.6     & -     & 83.4           \\
		OpenCUA-72B \citep{wang2026opencua}       & 60.8     & 37.3     & -           \\
		UI-Venus-72B \citep{gu2025uivenus}     & 61.9     & 36.8     & 86.3        \\
        Qwen3-VL-32B$^*$ \citep{bai2025qwen3vl} & 54.5  & 32.7     & 84.2     \\
		\rowcolor{gray!10}
		LookAgain-32B                     & 61.5  & \textbf{47.9}  & \textbf{88.4}  \\
		\bottomrule
	\end{tabular}
    \caption{
		\textbf{Overall results of LookAgain across three GUI-grounding benchmarks.}
		$^*$ denotes our evaluated results.
	}
    \label{tab:main_overview_results}
\end{table}

\subsection{Main Results}

\paragraph{Results on Refusal-Aware GUI Grounding} Table~\ref{tab:main_refusal_results} summarizes the results on OSWorld-G~\cite{xie2026osg} and VenusBench-GD~\cite{zhou2025venusbench}.
Overall, LookAgain achieves the best performance among models of comparable scale on both benchmarks, showing that the proposed multi-turn grounding paradigm substantially improves both standard grounding and refusal behavior.

On OSWorld-G~\cite{xie2026osg}, LookAgain-8B reaches 74.7 on Standard, 57.4 on Refusal, and 73.0 Overall, outperforming the base Qwen3-VL-8B~\cite{bai2025qwen3vl} by 21.7 points overall, respectively. Notably, many strong GUI models achieve competitive standard grounding accuracy but nearly fail on refusal cases. In contrast, LookAgain improves refusal performance substantially while still increasing standard grounding accuracy, indicating that the gains do not come from conservative prediction or over-refusal, but from better target verification. The same advantage carries over to the larger backbone, where LookAgain-32B lifts the base Qwen3-VL-32B~\cite{bai2025qwen3vl} by
+15.1 and attains the best overall accuracy. 

A similar trend is also observed on VenusBench-GD~\cite{zhou2025venusbench}, where LookAgain-8B achieves 69.8 overall score, outperforming Qwen3-VL-8B~\cite{bai2025qwen3vl} by 8.3 points. LookAgain-32B also gains 7.0 improvement. This result suggests that explicit re-checking before final localization is particularly effective in both general and refusal grounding tasks. \textbf{We also provide the results on OSWorld-G-Refine~\cite{xie2026osg} in Appendix.}

\paragraph{Results on General GUI Grounding} Table~\ref{tab:main_overview_results} further reports results on Screenspot-Pro~\cite{li2025screenspot}, UIVision~\cite{nayak2025ui}, and MMBench-L2-GUI~\cite{wang2025mmbench}. LookAgain-8B achieves 60.2 on Screenspot-Pro~\cite{li2025screenspot}, 38.8 on UIVision~\cite{nayak2025ui}, and 84.5 on MMBench-L2-GUI~\cite{wang2025mmbench}.
Compared with the base Qwen3-VL-8B model, it gains +13.1, +15.5, and +5.2 points increment, respectively. LookAgain-32B also improves Qwen3-VL-32B by +7.0, +15.2, and +4.2 points on the three benchmarks. The results indicate that LookAgain improves not only refusal handling, but also general-purpose grounding robustness.
Notably, these benchmarks all require fine-grained grounding under diverse layouts and substantial visual ambiguity. This demonstrates that the visually grounded reflection ability of LookAgain facilitates fine-grained element recognition across diverse categories.

\begin{figure}[t] 
    \centering
    \includegraphics[width=1\linewidth]{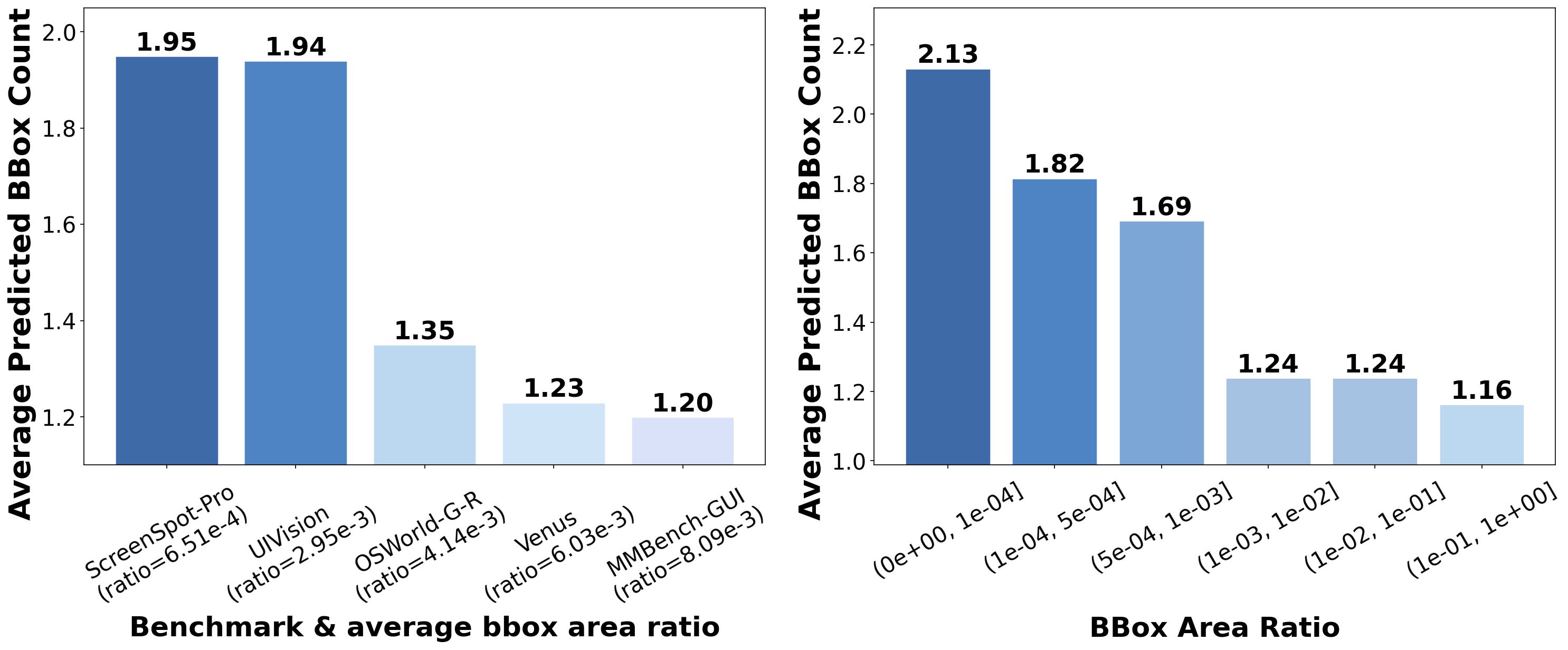}
    \caption{\textbf{Relationship between target size and prediction count in LookAgain.} The left figure shows the average number of predicted bounding boxes for each benchmark, together with the corresponding average target area ratio. The right figure shows the average number of predicted bounding boxes after grouping samples by target area ratio. The results suggest that LookAgain tends to perform more rounds of reflection for smaller targets, which helps collect finer-grained visual details for localization.}
    \label{fig:num_turn}
\end{figure}

\paragraph{Analysis on the prediction bounding box counts.} To better understand when LookAgain tends to perform more iterative grounding, we analyze the relationship between the relative target size and the average number of predicted bounding boxes. As shown in Figure~\ref{fig:num_turn}, there is a clear inverse relationship between the target-to-image area ratio and the average number of predicted bounding boxes, i.e. smaller targets generally lead to larger prediction counts. It is consistently observed both across benchmarks and across globally grouped samples by bbox area ratio. This indicates that LookAgain can effectively use more rounds to gather finer-grained visual details for reflection and verification when the target occupies a smaller portion of the image, which is particularly beneficial for small-object localization. We further provide more analysis on the prediction bbox counts in \textbf{Appendix}.

\begin{table}[!t]
	\centering
	\footnotesize
	\setlength{\tabcolsep}{1.5pt}
	\begin{tabular}{l c c c}
		\toprule
		\textbf{Model} & \textbf{SSPro} & \textbf{UIVision} & \textbf{\makecell{Venus-GD}} \\
		\midrule
		Qwen3-VL-8B$^*$ \citep{bai2025qwen3vl}  & 47.1 & 23.3 & 61.5 \\
        + \textit{GRPO-Single Turn w/o refusal}  & 55.7 & 31.9 & 66.6 \\
        + \textit{GRPO-Single Turn w/ refusal}  & 47.6 & 26.2 & 65.4 \\
		\rowcolor{gray!10}
		LookAgain-8B     & \textbf{60.2}  & \textbf{38.8} & \textbf{69.8} \\
		\bottomrule
	\end{tabular}
    \caption{\textbf{Comparison with single-turn grounding.} We compare the base model, a single-turn GRPO variant, and the full LookAgain model to analyze the effect of the proposed multi-turn look-again mechanism.}
    \label{tab:ablation_multiturn}
\end{table}

\begin{table}[!t]
	\centering
	\footnotesize
	\setlength{\tabcolsep}{5pt}
	\begin{tabular}{c c c c c}
		\toprule
		\textbf{SFT} & \textbf{GRPO} & \textbf{OSG-R} & \textbf{UIVision} & \textbf{\makecell{Venus-GD}} \\
		\midrule
		      &   & 67.8 & 23.3 & 61.5 \\
		\cmark &  & 73.5 & 37.5 & 60.6 \\
		\rowcolor{gray!10}
		\cmark  & \cmark  & \textbf{81.0}  & \textbf{38.8}     & \textbf{69.8} \\
		\bottomrule
	\end{tabular}
    \caption{\textbf{Pipeline ablation of LookAgain.} We study the contributions of SFT and GRPO in the full training pipeline.}
    \label{tab:ablation_pipeline}
\end{table}

\subsection{Ablation Study} 



\paragraph{Effect of multi-turn reflection.}
To validate that the performance gain mainly comes from the proposed multi-turn reflection mechanism rather than RL alone, we compare the base Qwen3-VL-8B, a GRPO-Single Turn variant, and the full LookAgain model in Table~\ref{tab:ablation_multiturn}.

Compared with the base model, single-turn GRPO brings only limited gains, e.g. improving performance from 47.1 to only 47.6 on SreenSpot-Pro. This marginal improvement suggest that directly training a single-turn policy to jointly handle standard grounding and refusal is suboptimal. In particular, learning to refuse within the same single-step decision process can interfere with standard grounding, increasing erroneous refusals on normal samples. In contrast, the LookAgain model boosts the performance to 60.2, substantially outperforming the single-turn variant. Even when the single-turn variant is relieved of refusal and reaches its best of 55.7 on SreenSpot-Pro, it still falls clearly short of LookAgain, showing that the gain stems from multi-turn reflection itself rather than from simply avoiding the refusal burden. We attribute this improvement to the multi-turn look-again procedure, which separates verification from immediate final prediction and therefore better balances accurate grounding with appropriate refusal.

\paragraph{Generality across base backbones.}
To verify the generalization ability of LookAgain beyond a single backbone, we apply the same strategy to Qwen3.5-9B~\cite{qwen35}. As shown in Table~\ref{tab:ablation_backbone}, LookAgain improves the performance of Qwen3.5-9B to 84.9 on OSWorld-G-R, 49.9 on UIVision, and 72.1 on Venus-GD. The large and consistent gains across all benchmarks suggest that LookAgain is not tightly coupled to a particular model family and can serve as a general enhancement recipe for GUI grounding models.

\paragraph{Pipeline ablation.}
To study the contribution of each training stage, we further conduct a pipeline ablation. As shown in Table~\ref{tab:ablation_pipeline}, adding only SFT improves OSWorld-G-Refine and UIVision to 73.5 and 37.5, but slightly decreases Venus-GD to 60.6, indicating that supervised tuning helps the model learn the interaction format and coarse grounding behavior, yet is insufficient for reliable refusal-aware decision making.
When GRPO is further added on top of the SFT model, performance rises sharply to 81.0, 38.8, and 69.8. We therefore conclude that the two stages are complementary. SFT provides a stable initialization for the multi-turn grounding process, while GRPO further strengthens verification, target discrimination, and refusal calibration.

\begin{table}[!t]
	\centering
	\footnotesize
	\setlength{\tabcolsep}{1.5pt}
	\begin{tabular}{l c c c}
		\toprule
		\textbf{Model} & \textbf{OSG-R} & \textbf{UIVision} & \textbf{\makecell{Venus-GD}} \\
		\midrule
		Qwen3.5-9B$^*$ \citep{qwen35}  & 74.6 & 34.5 & 59.2 \\
        + \textit{GRPO-Single Turn w/ refusal}  & 78.3 & 40.8 & 61.8 \\
		\rowcolor{gray!10}
		LookAgain-9B   & \textbf{84.9}  & \textbf{49.9} & \textbf{72.1} \\
		\bottomrule
	\end{tabular}
    \caption{\textbf{Backbone generalization of LookAgain.} The performance of LookAgain when instantiated on the Qwen3.5-9B backbone.}
    \label{tab:ablation_backbone}
\end{table}

%% file: Sections/sec5_conclusion.tex
\section{Conclusion}
To address the lack of post-hoc verification in GUI grounding, we propose LookAgain, a closed-loop grounding framework based on visually grounded reflection. LookAgain reformulates grounding as a multi-turn process of prediction, reflection and refinement, where \textit{locate} generates a prediction with a marker-annotated local patch, and \textit{confirm} commits the result. By grounding refinement in evidence elicited from previous hypothesis, LookAgain enables explicit reflection and correction before final commitment. We further introduce a two-stage training pipeline that cold-starts reflective grounding on constructed trajectories and optimizes terminal correctness through reinforcement learning. Extensive experiments show the superior performance of LookAgain.

%% file: aaai2027.bib
@inproceedings{cheng2024seeclick,
  title={Seeclick: Harnessing gui grounding for advanced visual gui agents},
  author={Cheng, Kanzhi and Sun, Qiushi and Chu, Yougang and Xu, Fangzhi and YanTao, Li and Zhang, Jianbing and Wu, Zhiyong},
  booktitle={Proceedings of the 62nd Annual Meeting of the Association for Computational Linguistics (Volume 1: Long Papers)},
  pages={9313--9332},
  year={2024}
}

@inproceedings{wu2025atlas,
  title={OS-ATLAS: Foundation action model for generalist GUI agents},
  author={Wu, Zhiyong and Wu, Zhenyu and Xu, Fangzhi and Wang, Yian and Sun, Qiushi and Jia, Chengyou and Cheng, Kanzhi and Ding, Zichen and Chen, Liheng and Liang, Paul Pu and others},
  booktitle={International Conference on Learning Representations},
  volume={2025},
  pages={5090--5108},
  year={2025}
}

@inproceedings{chen2026uiins,
  title={{UI}-Ins: Enhancing {GUI} Grounding with Multi-Perspective Instruction as Reasoning},
  author={Liangyu Chen and Hanzhang Zhou and chenglin cai and Jianan Zhang and Panrong Tong and Xu Zhang and Quyu Kong and Chen Liu and Yuqi Liu and Wenxuan Wang and Yue Wang and Qin Jin and Steven HOI},
  booktitle={The Fourteenth International Conference on Learning Representations},
  year={2026},
  url={https://openreview.net/forum?id=dsQHm7YX9c}
}

@article{luo2025gui,
  title={Gui-r1: A generalist r1-style vision-language action model for gui agents},
  author={Luo, Run and Wang, Lu and He, Wanwei and Chen, Longze and Li, Jiaming and Xia, Xiaobo},
  journal={arXiv preprint arXiv:2504.10458},
  year={2025}
}

@inproceedings{lu2026ui,
  title={Ui-r1: Enhancing efficient action prediction of gui agents by reinforcement learning},
  author={Lu, Zhengxi and Chai, Yuxiang and Guo, Yaxuan and Yin, Xi and Liu, Liang and Wang, Hao and Xiao, Han and Ren, Shuai and Zhao, Pengxiang and Liu, Guangyi and others},
  booktitle={Proceedings of the AAAI Conference on Artificial Intelligence},
  volume={40},
  number={21},
  pages={17608--17616},
  year={2026}
}

@inproceedings{tang2026guig2,
  title={GUI-G$^2$: Gaussian Reward Modeling for GUI Grounding},
  author={Tang, Fei and Gu, Zhangxuan and Lu, Zhengxi and Liu, Xuyang and Shen, Shuheng and Meng, Changhua and Wang, Wen and Zhang, Wenqi and Shen, Yongliang and Lu, Weiming and others},
  booktitle={Proceedings of the AAAI Conference on Artificial Intelligence},
  volume={40},
  number={39},
  pages={33214--33222},
  year={2026}
}

@article{zhou2026guig1,
  title={Gui-g1: Understanding r1-zero-like training for visual grounding in gui agents},
  author={Zhou, Yuqi and Dai, Sunhao and Wang, Shuai and Zhou, Kaiwen and Jia, Qinglin and Xu, Jun},
  journal={Advances in Neural Information Processing Systems},
  volume={38},
  pages={95683--95705},
  year={2026}
}

@article{wu2026guiactor,
  title={Gui-actor: Coordinate-free visual grounding for gui agents},
  author={Wu, Qianhui and Cheng, Kanzhi and Yang, Rui and Zhang, Chaoyun and Yang, Jianwei and Jiang, Huiqiang and Mu, Jian and Peng, Baolin and Qiao, Bo and Tan, Reuben and others},
  journal={Advances in Neural Information Processing Systems},
  volume={38},
  pages={15101--15128},
  year={2026}
}

@inproceedings{zhang2026mvp,
  title={Mvp: Multiple view prediction improves gui grounding},
  author={Zhang, Yunzhu and Pan, Zeyu and Zeng, Zhengwen and Shen, Shuheng and Meng, Changhua and Zhu, Linchao},
  booktitle={Proceedings of the IEEE/CVF Conference on Computer Vision and Pattern Recognition},
  pages={27482--27492},
  year={2026}
}

@inproceedings{luo2025regionfocus,
  title={Visual test-time scaling for gui agent grounding},
  author={Luo, Tiange and Logeswaran, Lajanugen and Johnson, Justin and Lee, Honglak},
  booktitle={Proceedings of the IEEE/CVF International Conference on Computer Vision},
  pages={19989--19998},
  year={2025}
}

@inproceedings{wu2025dimo,
  title={Dimo-gui: Advancing test-time scaling in gui grounding via modality-aware visual reasoning},
  author={Wu, Hang and Chen, Hongkai and Cai, Yujun and Liu, Chang and Ye, Qingwen and Yang, Ming-Hsuan and Wang, Yiwei},
  booktitle={Proceedings of the 2025 Conference on Empirical Methods in Natural Language Processing},
  pages={26257--26267},
  year={2025}
}

@article{shinn2023reflexion,
  title={Reflexion: Language agents with verbal reinforcement learning},
  author={Shinn, Noah and Cassano, Federico and Gopinath, Ashwin and Narasimhan, Karthik and Yao, Shunyu},
  journal={Advances in neural information processing systems},
  volume={36},
  pages={8634--8652},
  year={2023}
}

@inproceedings{gou2025navigating,
  title={Navigating the digital world as humans do: Universal visual grounding for gui agents},
  author={Gou, Boyu and Wang, Demi Ruohan and Zheng, Boyuan and Xie, Yanan and Chang, Cheng and Shu, Yiheng and Sun, Huan and Su, Yu},
  booktitle={International Conference on Learning Representations},
  volume={2025},
  pages={30851--30883},
  year={2025}
}

@inproceedings{yang2025aria,
  title={Aria-ui: Visual grounding for gui instructions},
  author={Yang, Yuhao and Wang, Yue and Li, Dongxu and Luo, Ziyang and Chen, Bei and Huang, Chao and Li, Junnan},
  booktitle={Findings of the Association for Computational Linguistics: ACL 2025},
  pages={22418--22433},
  year={2025}
}

@inproceedings{li2025screenspot,
  title={Screenspot-pro: Gui grounding for professional high-resolution computer use},
  author={Li, Kaixin and Meng, Ziyang and Lin, Hongzhan and Luo, Ziyang and Tian, Yuchen and Ma, Jing and Huang, Zhiyong and Chua, Tat-Seng},
  booktitle={Proceedings of the 33rd ACM International Conference on Multimedia},
  pages={8778--8786},
  year={2025}
}

@inproceedings{nayak2025ui,
  title={UI-Vision: A Desktop-centric GUI Benchmark for Visual Perception and Interaction},
  author={Nayak, Shravan and Jian, Xiangru and Lin, Kevin Qinghong and Rodriguez, Juan A and Kalsi, Montek and Chapados, Nicolas and {\"O}zsu, M Tamer and Agrawal, Aishwarya and Vazquez, David and Pal, Christopher and others},
  booktitle={International Conference on Machine Learning},
  pages={45817--45851},
  year={2025},
  organization={PMLR}
}

@article{liu2025infigui,
  title={Infigui-r1: Advancing multimodal gui agents from reactive actors to deliberative reasoners},
  author={Liu, Yuhang and Li, Pengxiang and Xie, Congkai and Hu, Xavier and Han, Xiaotian and Zhang, Shengyu and Yang, Hongxia and Wu, Fei},
  journal={arXiv preprint arXiv:2504.14239},
  year={2025}
}

@article{guo2025deepseek,
  title={Deepseek-r1: Incentivizing reasoning capability in llms via reinforcement learning},
  author={Guo, Daya and Yang, Dejian and Zhang, Haowei and Song, Junxiao and Wang, Peiyi and Zhu, Qihao and Xu, Runxin and Zhang, Ruoyu and Ma, Shirong and Bi, Xiao and others},
  journal={arXiv preprint arXiv:2501.12948},
  year={2025}
}

@article{chen2026v2p,
  title={V2P: Visual Attention Calibration for GUI Grounding via Background Suppression and Center Peaking},
  author={Chen, Jikai and Chen, Long and Wang, Dong and Su, Qinglin and Chu, Zhixuan and Hao, Bingguang and Gan, Leilei and Zhuang, Chenyi and Gu, Jinjie},
  journal={arXiv preprint arXiv:2601.06899},
  year={2026}
}

@article{bai2025qwen3vl,
  title={Qwen3-vl technical report},
  author={Bai, Shuai and Cai, Yuxuan and Chen, Ruizhe and Chen, Keqin and Chen, Xionghui and Cheng, Zesen and Deng, Lianghao and Ding, Wei and Gao, Chang and Ge, Chunjiang and others},
  journal={arXiv preprint arXiv:2511.21631},
  year={2025}
}

@article{yang2025gta1,
  title={Gta1: Gui test-time scaling agent},
  author={Yang, Yan and Li, Dongxu and Dai, Yutong and Yang, Yuhao and Luo, Ziyang and Zhao, Zirui and Hu, Zhiyuan and Huang, Junzhe and Saha, Amrita and Chen, Zeyuan and others},
  journal={arXiv preprint arXiv:2507.05791},
  year={2025}
}

@article{wang2026opencua,
  title={Opencua: Open foundations for computer-use agents},
  author={Wang, Xinyuan and Wang, Bowen and Lu, Dunjie and Yang, Junlin and Xie, Tianbao and Wang, Junli and Deng, Jiaqi and Guo, Xiaole and Xu, Yiheng and Wu, Chen and others},
  journal={Advances in Neural Information Processing Systems},
  volume={38},
  pages={139756--139806},
  year={2026}
}

@article{gu2025uivenus,
  title={Ui-venus technical report: Building high-performance ui agents with rft},
  author={Gu, Zhangxuan and Zeng, Zhengwen and Xu, Zhenyu and Zhou, Xingran and Shen, Shuheng and Liu, Yunfei and Zhou, Beitong and Meng, Changhua and Xia, Tianyu and Chen, Weizhi and others},
  journal={arXiv preprint arXiv:2508.10833},
  year={2025}
}

@misc{hai2025holo15modelfamily,
      title={Holo1.5 - Open Foundation Models for Computer Use Agents}, 
      author={H Company},
      year={2025},
      url={https://huggingface.co/collections/Hcompany/holo15-68c1a5736e8583a309d23d9b}, 
}

@article{qin2025uitars,
  title={Ui-tars: Pioneering automated gui interaction with native agents},
  author={Qin, Yujia and Ye, Yining and Fang, Junjie and Wang, Haoming and Liang, Shihao and Tian, Shizuo and Zhang, Junda and Li, Jiahao and Li, Yunxin and Huang, Shijue and others},
  journal={arXiv preprint arXiv:2501.12326},
  year={2025}
}

@misc{hai2025holo2modelfamily,
      title={Holo2 - Open Foundation Models for Navigation and Computer Use Agents}, 
      author={H Company},
      year={2025},
      url=https://huggingface.co/collections/Hcompany/holo2, 
}

@article{ye2025mobile,
  title={Mobile-agent-v3: Fundamental agents for gui automation},
  author={Ye, Jiabo and Zhang, Xi and Xu, Haiyang and Liu, Haowei and Wang, Junyang and Zhu, Zhaoqing and Zheng, Ziwei and Gao, Feiyu and Cao, Junjie and Lu, Zhengxi and others},
  journal={arXiv preprint arXiv:2508.15144},
  year={2025}
}

@misc{qwen35,
    title  = {{Qwen3.5}: Towards Native Multimodal Agents},
    author = {{Qwen Team}},
    month  = {February},
    year   = {2026},
    url    = {https://qwen.ai/blog?id=qwen3.5}
}

@article{xie2026osg,
  title={Scaling computer-use grounding via user interface decomposition and synthesis},
  author={Xie, Tianbao and Deng, Jiaqi and Li, Xiaochuan and Yang, Junlin and Wu, Haoyuan and Chen, Jixuan and Hu, Wenjing and Wang, Xinyuan and Xu, Yuhui and Wang, Zekun and others},
  journal={Advances in Neural Information Processing Systems},
  volume={38},
  year={2026}
}

@article{zhou2025venusbench,
  title={VenusBench-GD: A Comprehensive Multi-Platform GUI Benchmark for Diverse Grounding Tasks},
  author={Zhou, Beitong and Huang, Zhexiao and Guo, Yuan and Gu, Zhangxuan and Xia, Tianyu and Luo, Zichen and Tang, Fei and Kong, Dehan and Shang, Yanyi and Ou, Suling and others},
  journal={arXiv preprint arXiv:2512.16501},
  year={2025}
}

@article{wang2025mmbench,
  title={Mmbench-gui: Hierarchical multi-platform evaluation framework for gui agents},
  author={Wang, Xuehui and Wu, Zhenyu and Xie, JingJing and Ding, Zichen and Yang, Bowen and Li, Zehao and Liu, Zhaoyang and Li, Qingyun and Dong, Xuan and Chen, Zhe and others},
  journal={arXiv preprint arXiv:2507.19478},
  year={2025}
}

@article{qiu2026vista,
  title={VISTA: View-Consistent Self-Verified Training for GUI Grounding},
  author={Qiu, Xinyu and Zhang, Yunzhu and Jia, Heng and Shen, Shuheng and Meng, Changhua and Zhu, Linchao},
  journal={arXiv preprint arXiv:2606.14579},
  year={2026}
}

@article{zhao2025learning,
  title={Learning gui grounding with spatial reasoning from visual feedback},
  author={Zhao, Yu and Chen, Wei-Ning and Inan, Huseyin Atahan and Kessler, Samuel and Wang, Lu and Wutschitz, Lukas and Yang, Fangkai and Zhang, Chaoyun and Minervini, Pasquale and Rajmohan, Saravan and others},
  journal={arXiv preprint arXiv:2509.21552},
  year={2025}
}

@article{feizi2025groundcua,
  title={Grounding Computer Use Agents on Human Demonstrations},
  author={Feizi, Aarash and Nayak, Shravan and Jian, Xiangru and Lin, Kevin Qinghong and Li, Kaixin and Awal, Rabiul and L{\`u}, Xing Han and Obando-Ceron, Johan and Rodriguez, Juan A and Chapados, Nicolas and others},
  journal={arXiv preprint arXiv:2511.07332},
  year={2025}
}

@online{google_gemini_31_pro_preview,
  author  = {{Google}},
  title   = {Gemini 3.1 Pro Preview},
  year    = {2026},
  url     = {https://ai.google.dev/gemini-api/docs/models/gemini-3.1-pro-preview},
  urldate = {2026-07-21}
}

@inproceedings{li2021cutpaste,
  title={Cutpaste: Self-supervised learning for anomaly detection and localization},
  author={Li, Chun-Liang and Sohn, Kihyuk and Yoon, Jinsung and Pfister, Tomas},
  booktitle={Proceedings of the IEEE/CVF conference on computer vision and pattern recognition},
  pages={9664--9674},
  year={2021}
}

@inproceedings{wang2026measure,
  title={Measure Twice, Click Once: Co-evolving Proposer and Visual Critic via Reinforcement Learning for GUI Grounding},
  author={Wang, Wenkai and Li, Xiyun and Guo, Hongcan and Yu, Wenhao and Fang, Tianqing and Mi, Haitao and Yu, Dong and Zhang, Shengyu},
  booktitle={Proceedings of the 64th Annual Meeting of the Association for Computational Linguistics (Volume 1: Long Papers)},
  pages={20964--20984},
  year={2026}
}

@inproceedings{zhang2025falcon,
  title={Falcon: Resolving visual redundancy and fragmentation in high-resolution multimodal large language models via visual registers},
  author={Zhang, Renshan and Shao, Rui and Chen, Gongwei and Zhang, Miao and Zhou, Kaiwen and Guan, Weili and Nie, Liqiang},
  booktitle={2025 IEEE/CVF International Conference on Computer Vision (ICCV)},
  pages={23530--23540},
  year={2025},
  organization={IEEE}
}

@inproceedings{zhou2025hiconagent,
  author    = {Zhou, Xurui and Chen, Gongwei and Xie, Yuquan and Li, Zaijing and Zhou, Kaiwen and Wang, Shuai and Yang, Shuo and Tian, Zhuotao and Shao, Rui},
  title     = {Hiconagent: History context-aware policy optimization for gui agents},
  booktitle = {Proceedings of the IEEE/CVF Conference on Computer Vision and Pattern Recognition (CVPR)},
  year      = {2026},
}

@inproceedings{lyu2026personalalign,
  title={Personalalign: Hierarchical implicit intent alignment for personalized gui agent with long-term user-centric records},
  author={Lyu, Yibo and Chen, Gongwei and Shao, Rui and Guan, Weili and Nie, Liqiang},
  booktitle={Proceedings of the 64th Annual Meeting of the Association for Computational Linguistics (Volume 1: Long Papers)},
  pages={36074--36089},
  year={2026}
}

@inproceedings{chen2025less,
  title={Less is more: Empowering gui agent with context-aware simplification},
  author={Chen, Gongwei and Zhou, Xurui and Shao, Rui and Lyu, Yibo and Zhou, Kaiwen and Wang, Shuai and Li, Wentao and Li, Yinchuan and Qi, Zhongang and Nie, Liqiang},
  booktitle={2025 IEEE/CVF International Conference on Computer Vision (ICCV)},
  pages={5901--5911},
  year={2025},
  organization={IEEE}
}

@article{shao2024detecting,
  title={Detecting and grounding multi-modal media manipulation and beyond},
  author={Shao, Rui and Wu, Tianxing and Wu, Jianlong and Nie, Liqiang and Liu, Ziwei},
  journal={IEEE Transactions on Pattern Analysis and Machine Intelligence},
  year={2024},
}

@inproceedings{shao2023detecting,
  title={Detecting and grounding multi-modal media manipulation},
  author={Shao, Rui and Wu, Tianxing and Liu, Ziwei},
  booktitle={Proceedings of the IEEE/CVF Conference on Computer Vision and Pattern Recognition},
  pages={6904--6913},
  year={2023}
}

@inproceedings{shao2019multi,
  title={Multi-adversarial discriminative deep domain generalization for face presentation attack detection},
  author={Shao, Rui and Lan, Xiangyuan and Li, Jiawei and Yuen, Pong C},
  booktitle={Proceedings of the IEEE/CVF conference on computer vision and pattern recognition},
  pages={10023--10031},
  year={2019}
}

@article{li2026cogvla,
  title={CogVLA: Cognition-aligned vision-language-action models via instruction-driven routing \& sparsification},
  author={Li, Wei and Zhang, Renshan and Shao, Rui and He, Jie and Nie, Liqiang},
  journal={Advances in neural information processing systems},
  volume={38},
  pages={137646--137675},
  year={2026}
}

@inproceedings{li2026semanticvla,
  title={Semanticvla: Semantic-aligned sparsification and enhancement for efficient robotic manipulation},
  author={Li, Wei and Zhang, Renshan and Shao, Rui and Fang, Zhijian and Zhou, Kaiwen and Tian, Zhuotao and Nie, Liqiang},
  booktitle={Proceedings of the AAAI Conference on Artificial Intelligence},
  volume={40},
  number={22},
  pages={18397--18405},
  year={2026}
}

@inproceedings{hu2026behavior,
  title={From Abstraction to Instantiation: Learning Behavioral Representation for Vision-Language-Action Model},
  author={Hu, Bing and Li, Zaijing and Shao, Rui and Chen, Junda and Liu, April Hua and Zheng, WeiShi and Nie, Liqiang
},
  booktitle={International Conference on Learning Representations},
  year={2026}
}

@inproceedings{li2025lion,
  title={Lion-fs: Fast \& slow video-language thinker as online video assistant},
  author={Li, Wei and Hu, Bing and Shao, Rui and Shen, Leyang and Nie, Liqiang},
  booktitle={2025 IEEE/CVF Conference on Computer Vision and Pattern Recognition (CVPR)},
  pages={3240--3251},
  year={2025},
  organization={IEEE}
}

@article{li2025optimus,
  title={Optimus-3: Dual-router aligned mixture-of-experts agent with dual-granularity reasoning-aware policy optimization},
  author={Li, Zaijing and Xie, Yuquan and Shao, Rui and Chen, Gongwei and Guan, Weili and Jiang, Dongmei and Wang, Yaowei and Nie, Liqiang},
  journal={arXiv preprint arXiv:2506.10357},
  year={2025}
}
